%% file: main.tex
\documentclass[numbers,imaiai,webpdf]{ima-authoring-template}

\hypersetup{final, hypertexnames=false}

\usepackage[utf8]{inputenc}
\usepackage[T1]{fontenc}
\usepackage{booktabs}
\usepackage{amsfonts}
\usepackage{nicefrac}
\usepackage{microtype}
\usepackage{tikz}
\usepackage{bbm, xfrac}
\usepackage{parskip}

\newtheorem{theorem}{Theorem}
\newtheorem{lemma}{Lemma}
\theoremstyle{definition}
\newtheorem{definition}{Definition}

\newtheorem{example}{Example}
\newtheorem{method}{Method}
\newtheorem{remark}{Remark}
\newtheorem{result}{Corollary}

\newcommand{\dr}{\mathrm{d}}
\newcommand{\x}{\mathbf{x}}

\newcommand{\subref}[2]{\hyperref[#1]{\ref*{#1}#2}}

\title[Conditional validity of heteroskedastic conformal regression]{Conditional validity of heteroskedastic conformal regression}

\author{Nicolas Dewolf*, Bernard De Baets and Willem Waegeman
\address{\orgdiv{KERMIT, Department of Data Analysis and Mathematical Modelling}, \orgname{Ghent University}, \orgaddress{\country{Ghent, Belgium}}}}

\corresp[*]{Corresponding author: \href{email:nicolas.dewolf@ugent.be}{nicolas.dewolf@ugent.be}}
\authormark{Dewolf et al.}

\DOI{DOI HERE}
\copyrightyear{2024}
\vol{00}
\appnotes{Paper}
\copyrightstatement{Submitted to Information and Inference: A Journal of the IMA.}
\firstpage{1}

\begin{document}

\abstract{
    Conformal prediction, and split conformal prediction as a specific implementation, offer a distribution-free approach to estimating prediction intervals with statistical guarantees. Recent work has shown that split conformal prediction can produce state-of-the-art prediction intervals when focusing on marginal coverage, i.e.~on a calibration dataset the method produces on average prediction intervals that contain the ground truth with a predefined coverage level. However, such intervals are often not adaptive, which can be problematic for regression problems with heteroskedastic noise. This paper tries to shed new light on how prediction intervals can be constructed, using methods such as normalized and Mondrian conformal prediction, in such a way that they adapt to the heteroskedasticity of the underlying process. Theoretical and experimental results are presented in which these methods are compared in a systematic way. In particular, it is shown how the conditional validity of a chosen conformal predictor can be related to (implicit) assumptions about the data-generating distribution.
}

\maketitle

\input{Introduction}
\input{Problem}
\input{CP}
\input{Theory}
\input{Synthetic}
\input{Real}

\section{Discussion}

    This paper was motivated by the surge in interest in conformal prediction and accurate uncertainty quantification over the past few years. Although the benefits and validity of these methods have been illustrated on numerous data sets and in various domains, an important aspect remains less understood: the conditional performance. More specifically, conditioning on estimates of the residual variance, so as to make sure the models do not neglect the often underrepresented regions of high uncertainty, deserves more attention. In this paper, uncertainty-independence was studied in both a general setting, leading to the use of pivotal quantities, and in the case of explicit, parametric nonconformity measures. The latter allows to derive families of probability distributions for which conditional validity will hold whenever the instance space is subdivided based on the data noise, provided it can be expressed in terms of the chosen parameterization.

    \begin{figure}[H]
        \centering
        \includegraphics[width = .8\linewidth]{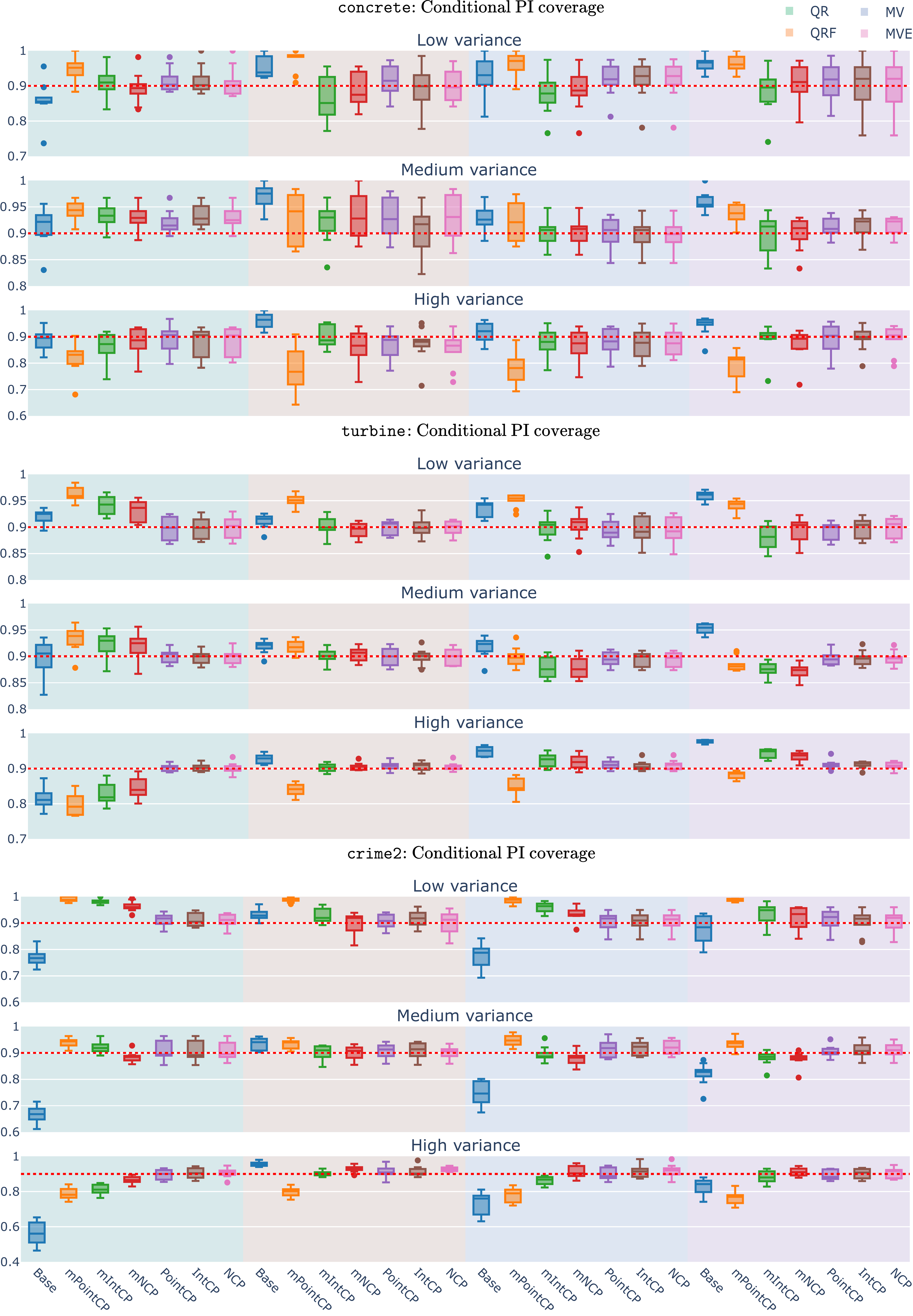}
        \caption{Conditional coverage at significance level $\alpha=0.1$ for the \texttt{concrete}, \texttt{turbine} and \texttt{crime2} data sets. The data is divided in three folds based on equal-frequency binning of the estimated variances. The coloured columns indicate the different estimators  (from left to right): quantile regression, quantile regression forest, mean-variance estimator and mean-variance ensemble. For every model, a baseline result and six nonconformity measures are shown (from left to right): residual, interval and $\widehat{\sigma}$-normalized nonconformity measures and their Mondrian counterparts.}
        \label{fig:cov_conditional}
    \end{figure}

    \begin{figure}[H]
        \centering
        \includegraphics[width = .8\linewidth]{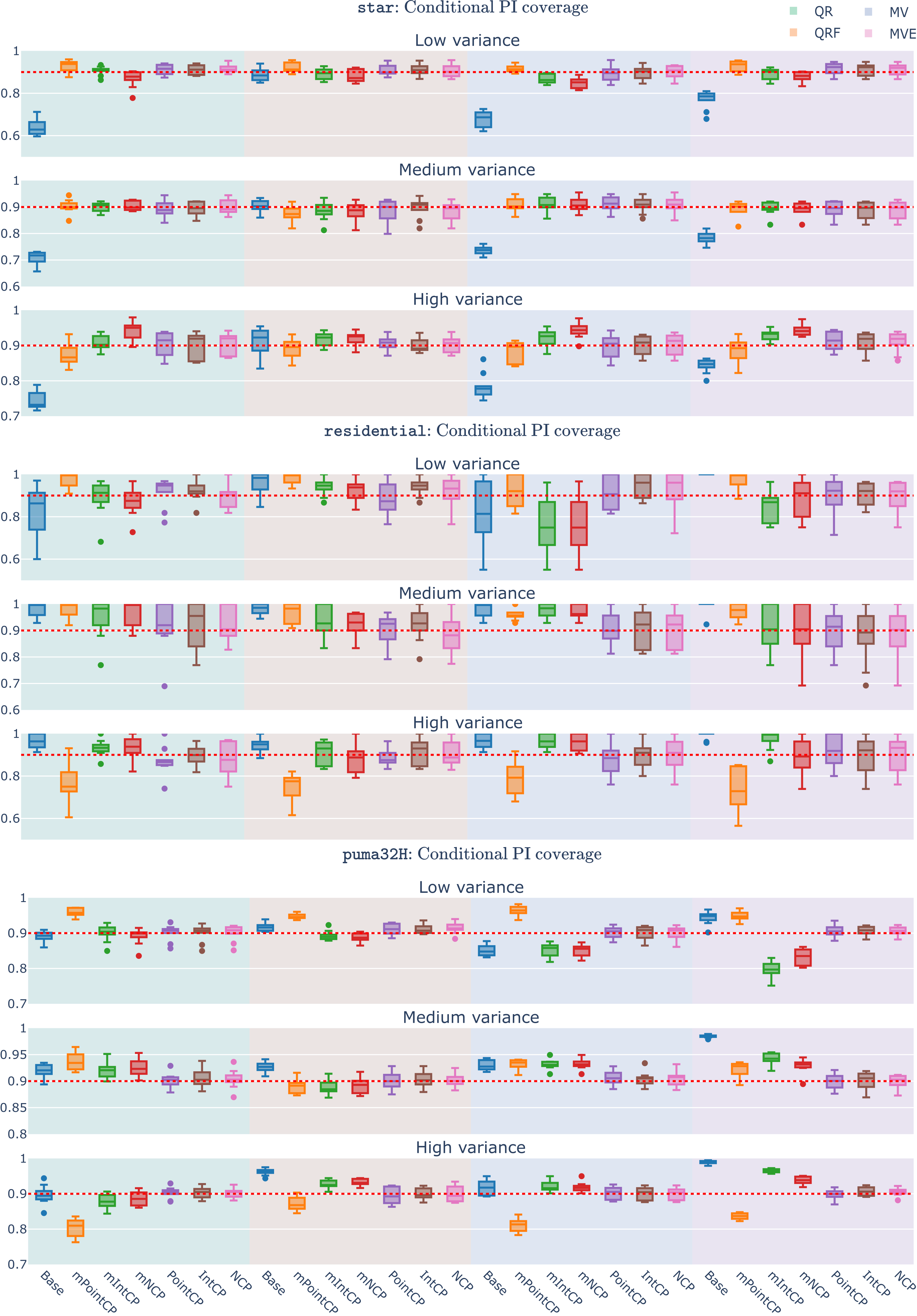}
        \caption{Conditional coverage at significance level $\alpha=0.1$ for the \texttt{star}, \texttt{residential} and \texttt{puma32H} data sets. The data is divided in three folds based on equal-frequency binning of the estimated variances. The coloured columns indicate the different estimators  (from left to right): quantile regression, quantile regression forest, mean-variance estimator and mean-variance ensemble. For every model, a baseline result and six nonconformity measures are shown (from left to right): residual, interval and $\widehat{\sigma}$-normalized nonconformity measures and their Mondrian counterparts.}
        \label{fig:cov_conditional2}
    \end{figure}

\bibliographystyle{abbrvnat}
\bibliography{Biblio}

\appendix
\input{Appendix}

\section*{Data availability}

    The code underlying this article is available on GitHub, at \url{https://github.com/nmdwolf/HeteroskedasticConformalRegression}.

\section*{Funding}

    This work was supported by the Flemish Government under the ``Onderzoeksprogramma Artifici\"ele Intelligentie (AI) Vlaanderen'' programme.

\end{document}

%% file: Introduction.tex
\section{Introduction}

\begin{figure}[t!]
        \centering
        \includegraphics[width = \linewidth]{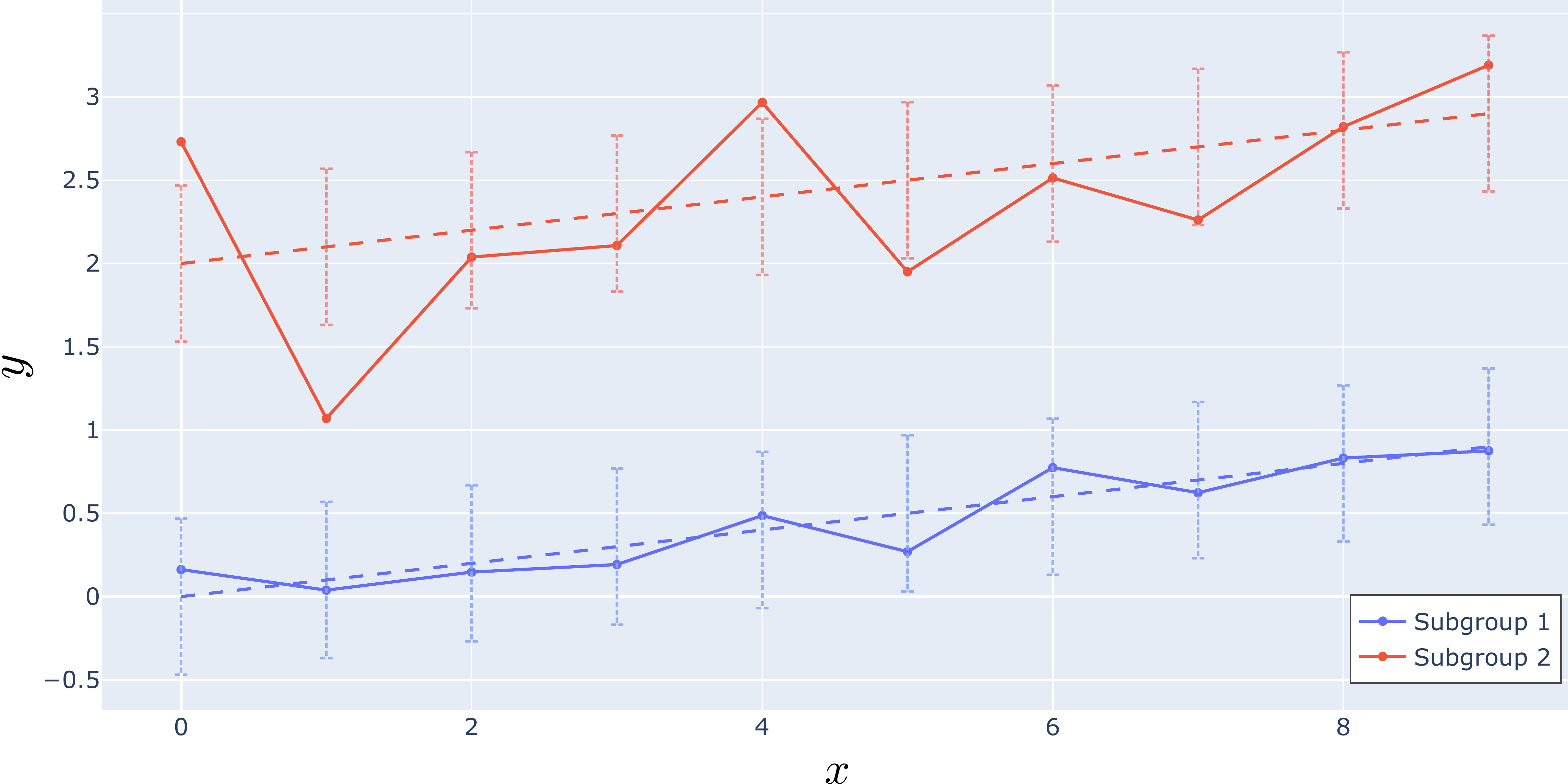}
        \caption{Two data samples with the same trend but with different noise levels: $y(x,s)\sim0.1x+2s+\varepsilon(s)$, where $s\in\{0,1\}$ is a dummy variable labelling the subgroups. The blue subgroup ($s=0$) has standard deviation $0.1$, while the red subgroup ($s=1$) has standard deviation $0.5$. Although the prediction intervals are valid at the $\alpha=0.2$ significance level, both marginally and for the blue subgroup, this is not the case for the red subgroup.}
        \label{fig:abstract}
    \end{figure}

    Many methods exist to estimate prediction sets or, more specifically, prediction intervals in the regression setting. Examples include Gaussian processes~\cite{williams1996gaussian}, quantile regression~\cite{koenker2001quantile} and Monte Carlo Dropout~\cite{srivastava2014dropout}. In a model-independent and distribution-free way, conformal prediction~\cite{angelopoulos-gentle,algorithmic} allows to estimate such regions with statistical guarantees. Different approaches to conformal prediction exist, e.g.~transductive~\cite{saunders1999transduction}, inductive~\cite{papadopoulos2002inductive,cqr} and cross-conformal prediction~\cite{vovk2015cross}. The validity of inductive conformal prediction has been verified numerous times, see e.g.~\cite{bosc2019large,toccaceli2017conformal,zhang2020inductive}, and a comparison of the aforementioned uncertainty quantification methods and further improvements resulting from applying conformal prediction as a (post-hoc) calibration method was carried out by the present authors~\cite{dewolf2023valid}. However, this analysis was only performed with respect to the entire, marginal data-generating distribution, where no substructure inherent to the data and problem setting is taken into account. Nonetheless, an important problem in the field of uncertainty quantification is exactly this conditional behaviour. Conformal prediction has a probabilistic validity guarantee, but this only holds w.r.t.~the full data distribution, i.e.~on average over the whole instance space. Consequently, the algorithm is allowed to attain the claimed validity by solely focusing on the `easy' parts of the data, which are often more abundant, while ignoring the more difficult parts. In practice it are, however, usually these difficult regions that matter the most. In this regard, consider Fig.~\ref{fig:abstract}. If the two samples considered would make up a data set, the prediction intervals would be valid at the significance level $\alpha=0.2$, because 80\% of the points is covered. Since these intervals were generated with a standard conformal predictor based on the absolute residuals for the significance level $\alpha=0.2$, this figure illustrates its marginal guarantees. However, all of the data points in the blue subgroup are covered, while only 60\% of the red subgroup is covered. The conformal predictor might work marginally as promised, but it is definitely not sufficient when working with data sets in which more structure is present.

    With the rise of conformal prediction, the interest in distribution-free conditional uncertainty modelling has also increased. Although Venn and Mondrian conformal predictors are actually almost as old as the field itself~\cite{algorithmic,vovk2003mondrian}, adoption by mainstream machine learning practitioners has remained even more limited than is the case for their nonconditional counterparts. Just like these counterparts, the conditional variants provide strict statistical validity guarantees, but this leads to an inherent problem when conditioning on sets of probability zero~\cite{foygel2021limits,vovk2012conditional}. In general it is not possible to obtain distribution-free guarantees for object-level conditioning, i.e.~when conditioning on the feature tuple. This forces researchers to aggregate data into larger subsets, thereby potentially reintroducing the issue of neglecting underrepresented regions of the instance space.

    A key concept in statistics is heteroskedasticity, where the conditional distributions for different values of the conditioning variable have a different variance. In this paper, the focus lies on modelling heteroskedastic noise with guarantees conditional on the level of heteroskedasticity, i.e.~where the data set is divided based on an estimate of the residual variance, as in Fig.~\ref{fig:abstract}, and the validity of different models w.r.t.~such a division is investigated. In this respect, it can be seen as a continuation of Bostr\"om \textit{et al.}~\cite{pmlr-v128-bostrom20a}. Aside from comparing the conditional validity of various standard nonconformity measures, with and without Mondrian taxonomies, theoretical conditions are derived for attaining conditional validity, with normalized conformal prediction as the prototypical example.

    In Section~\ref{section:problem}, the problem setting is discussed in more detail and a formal definition of (conditional) validity is given. Section~\ref{section:cp} covers the general framework of (inductive) conformal prediction and how it can be applied in a conditional context. Both the use of normalized nonconformity measures and Mondrian conformal predictors is covered. The explicit case of uncertainty-dependent conditioning is treated in Section~\ref{section:uncertainty} in a theoretical way, highlighting how various common parametric models give rise to conditionally valid (normalized) conformal predictors in a natural way. Before considering some real-world data sets in Section~\ref{section:experiment}, the impact of misspecification is analyzed in Section~\ref{section:synthetic}, in which some practical diagnostic tools are introduced that can help data scientists to decide on which framework to use.

%% file: Problem.tex
\section{Problem statement}\label{section:problem}

    As mentioned in the introduction, the main focus of this paper lies on conditional uncertainty quantification and, in particular, the construction of prediction regions with conditional guarantees. This conditioning is induced by a subdivision of the instance space, which in turn is performed using a so-called \textit{taxonomy function}~\cite{algorithmic,vovk2012conditional}. To formalize this problem, some notations and conventions are fixed. Firstly, it is worth mentioning that some abuse of notation and terminology will be present, e.g.~sets and multisets will be treated on an equal footing, as will unions and disjoint unions. Moreover, wherever necessary, functions will be assumed to be measurable.
    
    In conformal prediction (to be introduced in Section~\ref{section:cp}) and, by extension, all of statistics and data science, the natural setting is that of data sequences $\bigl(\!(\x_n,y_n)\!\bigr)_{n\in\mathbb{N}}$ in $\mathcal{X}\times\mathbb{R}$, where the target space has been fixed to $\mathbb{R}$ since only (univariate) regression problems are of interest in this paper (note that most of the definitions in this and the ensuing sections can be generalized to arbitrary target spaces):
    \begin{gather}
        \label{point_predictor}
        y_i = \widehat{y}(\mathbf{x}_i) + \varepsilon_i\,.
    \end{gather}
    The set of all sequences in $\mathcal{X}\times\mathbb{R}$ will be denoted by $(\mathcal{X}\times\mathbb{R})^\infty:=\cup_{n=1}^\infty(\mathcal{X}\times\mathbb{R})^n$. The feature space $\mathcal{X}$ can be any type of space, such as $\mathbb{R}^n$, $\mathbb{N}$, etc. Elements of this space are denoted by bold font symbols: $\x\in\mathcal{X}$. Although in machine learning the data sequence $\bigl(\!(\x_n,y_n)\!\bigr)_{n\in\mathbb{N}}$ is often assumed to be drawn identically and independently from a joint distribution $P_{X,Y}$, conformal prediction relaxes this requirement~\cite{algorithmic} to $\bigl(\!(\x_n,y_n)\!\bigr)_{n\in\mathbb{N}}$ being drawn \textit{exchangeably} from $P_{X,Y}$. Cumulative distributions will be denoted by the capital letter $F$ and, if they exist, probability density functions will be denoted by a lower case $f$. For clarity, all estimators will be denoted by a caret, e.g.~$\widehat{\mu}$ and $\widehat{\sigma}$ denote estimators of the (conditional) mean $\mu$ and (conditional) standard deviation $\sigma$ of $P_{Y\mid X}$, respectively.

    \begin{definition}[Validity]\label{validity}
        An interval predictor $\Gamma^\alpha:\mathcal{X}\rightarrow[\mathbb{R}]$ is said to be \textit{(marginally) valid} at significance level $\alpha\in[0,1]$ if
        \begin{gather}
            \mathrm{Prob}\bigl(Y\in\Gamma^\alpha(X)\bigr)\geq1-\alpha\,,
        \end{gather}
        where $[\mathbb{R}]$ denotes the set of all (closed) intervals in $\mathbb{R}$:
        \begin{gather}
            [\mathbb{R}]:=\bigl\{[a,b]\,\big\vert\,a,b\in\mathbb{R}\land a\leq b\bigr\}\,.
        \end{gather}
        Consider a function $\kappa:\mathcal{X}\times\mathbb{R}\rightarrow\mathcal{C}$, called the \textit{taxonomy function}. The interval predictor $\Gamma^\alpha:\mathcal{X}\rightarrow[\mathbb{R}]$ is said to be \textit{conditionally valid} w.r.t.~$\kappa$ at significance level $\alpha\in[0,1]$ if
        \begin{gather}
            \mathrm{Prob}\bigl(Y\in\Gamma^\alpha(X)\,\big\vert\,\kappa(X,Y)=c\bigr)\geq1-\alpha
        \end{gather}
        for all $c\in\mathcal{C}$.
    \end{definition}

    For convenience, the taxonomy space $\mathcal{C}$ and the distribution $P_C$ of the taxonomy class $C:=\kappa(X,Y)$, given by the pushforward rule
    \begin{gather}
        \label{taxonomy_distribution}
        P_C(c) := P_{X,Y}\bigl(\kappa^{-1}(c)\bigr)\,,
    \end{gather}
    where the preimage is defined by
    \begin{gather}
        \kappa^{-1}(c) := \bigl\{(\x,y)\in\mathcal{V}\bigm\vert\kappa(\x,y)=c\bigr\}\,,
    \end{gather}
    are assumed to be discrete with only the empty set having probability zero, such that conditioning on a taxonomy class does not lead to measure-zero issues. Note that the taxonomy function $\kappa$ can, in general, be any function. However, for the purpose of this paper, a specific type of taxonomy will be considered. The taxonomy functions of interest divide the instance space based on an estimate of the uncertainty \cite{pmlr-v128-bostrom20a}. In this paper, the taxonomy function will be derived from a proxy of the heteroskedastic noise such as the (conditional) standard deviation. More formally, given such an estimate $\delta:\mathcal{X}\rightarrow\mathbb{R}^+$ and a binning function $\mathcal{B}:\mathbb{R}^+\rightarrow\mathcal{C}$, the induced taxonomy function is given by
    \begin{gather}
        \label{taxonomy_function}
        \kappa:=\mathcal{B}\circ\delta\circ\pi_1:\mathcal{X}\times\mathbb{R}\rightarrow\mathcal{C}\,,
    \end{gather}
    where $\pi_1:\mathcal{X}\times\mathbb{R}\rightarrow\mathcal{X}:(\x,y)\mapsto\x$ projects a data point onto its features. A straightforward choice would be where $\delta=\widehat{\sigma}$ is an estimate for the (conditional) standard deviation and $\mathcal{B}$ corresponds to equal frequency binning for some predetermined number of classes (see, e.g.,~Fig.~\ref{fig:variance_cdf} further below for the case of three classes).

\begin{figure}[t]
        \centering
        \includegraphics[width = .99\linewidth]{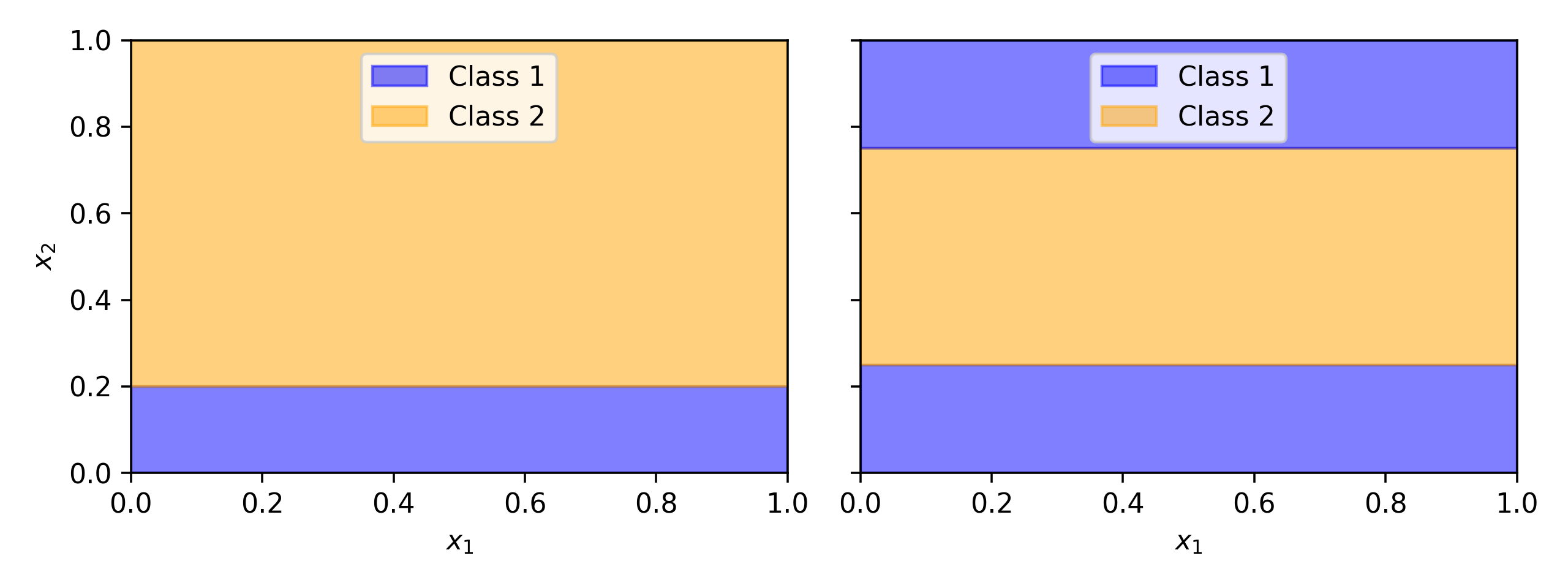}
        \caption{Division of the feature space based on two different taxonomy functions. The first one simply thresholds the second dimension (at $\xi=0.2$), whereas the second one performs equal-frequency binning on the (conditional) standard deviation, which has the form shown in~Eq.~\eqref{problem_example_distribution}.}
        \label{fig:illustration}
    \end{figure}

    For clarity's sake, a simple example is in order.
    \begin{example}\label{problem_example}
        Consider the two-dimensional feature space $\mathcal{X}=[0,1]^2$, equipped with the uniform distribution $\mathcal{U}^2$. As data-generating process, take\footnote{The reason for this choice will be explained in Section~\ref{section:synthetic}, where this example will be studied in more detail.}
        \begin{gather}
            \label{problem_example_distribution}
            y(\x)\sim\mathcal{N}\bigl(x_1+x_2,1+|x_2-0.5|\bigr)\,.
        \end{gather} 
        One possible choice of taxonomy function is one that divides the instance space based on some parameters such as
        \begin{gather}
            \kappa_\xi(\x,y) = \mathbbm{1}_{[0,\xi]}(x_2)\,,
        \end{gather}
        where $\xi\in\mathbb{R}$ and 
        \begin{gather}
            \label{indicator_function}
            \mathbbm{1}_S(x) :=
            \begin{cases}
                1 & \mbox{if } x\in S\,,\\
                0 & \mbox{if } x\not\in S\,,
            \end{cases}
        \end{gather}
        denotes the indicator function of the set $S$. The resulting division of the feature space is shown in the left-hand side of Fig.~\ref{fig:illustration}. Another possibility, giving an example of the general class of taxonomies defined by Eq.~\eqref{taxonomy_function}, would be to divide the feature space by binning the (conditional) standard deviation. For the case of two classes, this is shown in the right-hand side of Fig.~\ref{fig:illustration}. Note that the former, feature-dependent taxonomy is entirely independent of the conditional distribution $P_{Y\mid X}$, while the division of the instance space by the latter, uncertainty-dependent taxonomy is strongly influenced by the form of $P_{Y\mid X}$.
    \end{example}

%% file: CP.tex
\section{Conformal prediction}\label{section:cp}

    Since conformal prediction \cite{angelopoulos-gentle,algorithmic} is the main framework used in this paper, a short introduction is in order. For simplicity and computational ease-of-use, attention is restricted to inductive (or split) conformal prediction (ICP) \cite{papadopoulos2002inductive}, where a data splitting strategy is adopted to avoid retraining the models, thereby sacrificing some statistical power.

\subsection{Inductive conformal regression}

    Everything starts with a choice of \textit{nonconformity measure}
    \begin{gather}
        A:\mathcal{X}\times\mathbb{R}\rightarrow\mathbb{R}\,,
    \end{gather}
    i.e.~a function assigning to every data point a nonconformity score, indicating how `weird' or `nonconform' it is. Since this function often depends on a training set $\mathcal{T}$, it can be interpreted as the weirdness w.r.t.~that data set. Similar to Eq.~\eqref{taxonomy_distribution}, the induced distribution of nonconformity scores is given by
    \begin{gather}
        P_A(B) := P_{X,Y}\bigl(A^{-1}(B)\bigr)\,,
    \end{gather}
    for all subsets $B\subseteq\mathbb{R}$. 
    
    Given a choice of nonconformity measure, the \textit{inductive} (or \textit{split}) \textit{conformal prediction} algorithm can be summarized as follows:
    \begin{enumerate}
        \itemindent=10pt
        \item (Optional) Choose a training set $\mathcal{T}\in(\mathcal{X}\times\mathbb{R})^\infty$ and train the underlying model of $A$.
        \item Choose a calibration set $\mathcal{V}\in(\mathcal{X}\times\mathbb{R})^\infty$ and significance level $\alpha\in[0,1]$.
        \item For every calibration point $(\mathbf{x}_i,y_i)\in\mathcal{V}$, calculate the nonconformity score $a_i:=A(\mathbf{x}_i,y_i)$.
        \item Calculate the \textit{critical nonconformity score}:
        \begin{gather}
            a^*_\mathcal{V} := q_{(1-\alpha)(1+\sfrac{1}{|\mathcal{V}|})}\bigl(\bigl\{A(\mathbf{x},y)\,\big\vert\,(\mathbf{x},y)\in\mathcal{V}\bigr\}\bigr)\,,
        \end{gather}
        where $q_\beta$ denotes the $\beta$-quantile with $\beta\in[0,1]$.
        \item For every new data point $\mathbf{x}\in\mathcal{X}$, include all elements $y\in\mathbb{R}$ in $\Gamma^\alpha(\mathbf{x})$ for which $A(\mathbf{x},y)\leq a^*_\mathcal{V}$.
    \end{enumerate}
    The reason for the `inflated' quantile in the definition of the critical score can be found in the theorem at the end of this introduction. It is used to correct for not knowing the true label of new data points. Pseudocode for the algorithm is given in Algorithm~\ref{algorithm:icp}.

    \begin{algorithm}[t!]
        \caption{Inductive Conformal Prediction}
        \label{algorithm:icp}
        \begin{algorithmic}[1]
            \Procedure{ICP}{$A,\alpha,\mathcal{T},\mathcal{V}$}
            \State (Optional) Train the underlying model of $A$ on $\mathcal{T}$
            \For{$(\mathbf{x}_i,y_i)\in\mathcal{V}$}
                \State Apply $A$: $a_i\leftarrow A(\mathbf{x}_i,y_i)$
            \EndFor
            \State Determine the critical value of $A$: $a^*\leftarrow\bigl((1-\alpha)(1 + \frac{1}{|\mathcal{V}|})\bigr)\text{-quantile of }\bigl\{a_i\mid(\mathbf{x}_i,y_i)\in\mathcal{V}\bigr\}$
            \State Construct an interval estimator $\Gamma^\alpha$ as follows:
            \Procedure{$\Gamma^\alpha$}{$\mathbf{x}\in\mathcal{X}$}
                \State\Return $\bigl\{y\in\mathbb{R}\mid A(\mathbf{x},y)\leq a^*\bigr\}$
            \EndProcedure
            \State\Return{$\Gamma^\alpha$}
            \EndProcedure
        \end{algorithmic}
    \end{algorithm}

    In general, conformal prediction allows for any choice of predictive model or nonconformity measure. However, even though for a fixed model the choice of nonconformity measure is virtually unconstrained, some choices are much more natural than others. The most widely used ones are the (absolute) \textit{residual measure}
    \begin{gather}
        \label{residual_score}
        A_\text{res}(\mathbf{x},y) := |\widehat{y}(\mathbf{x})-y|
    \end{gather}
    in the case of point predictors~\cite{algorithmic} as in~\eqref{point_predictor}, and the \textit{interval measure}
    \begin{gather}
        \label{interval_score}
        A_\text{int}(\mathbf{x},y) := \max\bigl(y - \widehat{y}_+(\mathbf{x}),\widehat{y}_-(\mathbf{x})-y\bigr)
    \end{gather}
    in the case of interval predictors \cite{cqr}, where $\widehat{y}_\pm:\mathcal{X}\rightarrow\mathbb{R}$ denote the upper and lower bound of the prediction intervals, respectively. In this paper, two other common approaches are also considered: \textit{normalized conformal prediction} (in the next section) and \textit{Mondrian conformal prediction} (in Section~\ref{section:mcp}). Both of these methods use estimates of the heteroskedastic noise, the function $\delta:\mathcal{X}\rightarrow\mathbb{R}^+$, in an explicit way.

    The power of all these (inductive) conformal prediction methods lies in the following theorem \cite{cqr,algorithmic}, where the notion of interval predictors is generalized to functions of the form
    \begin{gather}
        \Gamma^\alpha:\mathcal{X}\times(\mathcal{X}\times\mathbb{R})^\infty\rightarrow[\mathbb{R}]
    \end{gather}
    as to make the dependence on the calibration set more apparent.
    \begin{theorem}[Marginal validity]\label{marginal_validity}
        Let $\Gamma^\alpha:\mathcal{X}\times(\mathcal{X}\times\mathbb{R})^\infty\rightarrow[\mathbb{R}]$ be an inductive conformal predictor at significance level $\alpha\in[0,1]$. If the nonconformity scores are exchangeable for any calibration set $\mathcal{V}$ and any new observation $(\mathbf{x},y)$, i.e.~any ordering of $A(\mathcal{V})\cup\{A(\mathbf{x},y)\}$ is equally probable, then $\Gamma^\alpha$ is conservatively valid:
        \begin{gather}
        \mathrm{Prob}\bigl(Y\in\Gamma^\alpha(X,V)\bigr)\geq1-\alpha\,,
        \end{gather}
        where the probability is taken over both $(X,Y)$ and $V$. Moreover, if the nonconformity scores are almost surely distinct, the conformal predictor is asymptotically exactly valid:
        \begin{gather}
            \mathrm{Prob}\bigl(Y\in\Gamma^\alpha(X,V)\,\big\vert\,|V|=n\bigr)\leq1-\alpha + \frac{1}{n+1}\,.
        \end{gather}
    \end{theorem}
    This theorem heavily relies on the following lemma \cite{cqr}.
    
    \begin{lemma}
        Let $\{X_1,\ldots,X_{n+1}\}$ be a set of exchangeable random variables for some $n\in\mathbb{N}_0$. The following relation holds for any $\alpha\in [0,1]$:
        \begin{gather}
            \mathrm{Prob}\left(X_{n+1}\leq\widehat{Q}_n\left(\!(1-\alpha)\left(1 + \frac{1}{n}\right)\!\right)\!\!\right)\geq1-\alpha\,,
        \end{gather}
        where $\widehat{Q}_n$ is the empirical quantile function of the set $\{X_1,\ldots,X_n\}$. Moreover, if ties almost surely do not arise, then this probability is also bounded from above by $1-\alpha+\frac{1}{n+1}$.
    \end{lemma}
    
    Note that replacing the (inflated) sample quantile by the true quantile $q_{1-\alpha}$ of the nonconformity distribution $P_A$ would give the same result. In practice this is how (inductive) conformal prediction is applied, unless suitable modifications, such as on-line training, are utilized~\cite{algorithmic}. It is assumed that the sample quantile is a consistent estimator, i.e.~converges in probability to $q_{1-\alpha}$. This is, for example, the case when the data is i.i.d.~and $P_A$ has unique quantiles.
    Instead of resampling a calibration set for every new test point, a fixed calibration set is used and the hoped-for consistency is assumed, allowing for violations of the above theorem. Therefore, from here on, the dependence on the calibration set will be left implicit.

\subsection{Normalized conformal prediction (NCP)}

    The standard residual measure~\eqref{residual_score} does not take into account any information about subregions of $\mathcal{X}$, such as where the model might perform subpar. As a consequence, the resulting prediction intervals are all of the same size, given by twice the critical value $a^*_\mathcal{V}$. As such, this method assumes domain knowledge about the homoskedasticity of the problem. To resolve this issue, knowledge about the data noise can be explicitly incorporated to obtain more realistic and more efficient intervals (meaning that the intervals will be smaller when possible). This leads to the idea of normalized conformal prediction~\cite{johansson2021investigating,papadopoulos2008normalized}:
    \begin{gather}
        \label{normalized_score}
        A_\delta(\mathbf{x},y) := \frac{|\widehat{y}(\mathbf{x})-y|}{\delta(\mathbf{x})}\,,
    \end{gather}
    where $\delta:\mathcal{X}\rightarrow\mathbb{R}^+$ is called the \textit{difficulty function} (suggestively given the same notation as the uncertainty measure inducing uncertainty-dependent taxonomies from the previous section). As before, in practice, $\delta$ will often be a characteristic of the heteroskedastic noise such as (an estimate of) the standard deviation.

    \begin{example}[Mean-variance estimators]\label{mve}
        A straightforward example of normalized nonconformity measures occurs in the case of mean-variance estimators~\cite{nix1994estimating}. Instead of simply estimating a point-predictor as in Eq.~\eqref{point_predictor}, this approach assumes a parametric model for the data-generating process, usually a normal distribution, characterized by a conditional mean $\widehat{\mu}:\mathcal{X}\rightarrow\mathbb{R}$ and conditional standard deviation $\widehat{\sigma}:\mathcal{X}\rightarrow\mathbb{R}^+$. These functions are then estimated through maximum likelihood estimation. In this case, the canonical choice of nonconformity score is\footnote{Note that for conformal prediction to work, the data does not have to be generated by some parametric model. The estimates $\widehat{\mu}$ and $\widehat{\sigma}$ can be obtained by means of any functions. However, assigning a (valid) probabilistic interpretation to these functions rests on the choice of parametric model. Moreover, the importance of parametric assumptions will be emphasized throughout the remainder of this paper.}
        \begin{gather}
            \label{ncp_score}
            A_{\widehat{\sigma}}(\mathbf{x},y)=\frac{|\widehat{\mu}(\mathbf{x})-y|}{\widehat{\sigma}(\mathbf{x})}\,.
        \end{gather}
        For stability issues \cite{johansson2021investigating}, $\widehat{\sigma}$ can be replaced by $\widehat{\sigma}+\varepsilon$ for some (small) $\varepsilon\in\mathbb{R}^+$. Note that this transformation induces a strong bias when $\varepsilon$ is not carefully tuned, especially when the variance is small compared to $\varepsilon$.

        Given a mean-variance estimator $(\widehat{\mu},\widehat{\sigma}):\mathcal{X}\rightarrow\mathbb{R}\times\mathbb{R}^+$, a prediction interval at significance level $\alpha\in[0,1]$ can be obtained by, for example, assuming that the data-generating distribution is a normal distribution. This gives rise to the following parametric form:
        \begin{gather}
            \label{mve_interval}
            \Gamma^\alpha_\text{MV}(\x) := \bigl[\widehat{\mu}(\x)-z^\alpha\widehat{\sigma}(\x),\widehat{\mu}(\x)+z^\alpha\widehat{\sigma}(\x)\bigr]\,,
        \end{gather}
        where $z^\alpha$ is the $(1-\alpha/2)$-quantile of the standard normal distribution $\mathcal{N}(0,1)$.
    \end{example}
    
    \begin{remark}
        One could argue that the quantiles of a Student $t$-distribution should be used because the standard deviation is merely an estimate. However, since the use of this interval is already based on a strong normality assumption and, moreover, the data sets in practice are quite large, the influence of this additional approximation should be minimal.
    \end{remark}

\subsection{Mondrian conformal predictor (MCP)}\label{section:mcp}

    Another possibility to explicitly incorporate the noise is given by Mondrian conformal prediction~\cite{vovk2003mondrian,vovk2012conditional}. Here, the data set is, just like the instance space, divided into multiple classes using the taxonomy function $\kappa:\mathcal{X}\times\mathbb{R}\rightarrow\mathcal{C}$. To this end, $\mathcal{C}$ is from here on also assumed to be finite such that the algorithm is numerically feasible in practice. This partitioning also induces a partitioning of the calibration set:
    \begin{gather}
        \mathcal{V}=\bigcup_{c\in\mathcal{C}}\mathcal{V}_c
    \end{gather}
    with
    \begin{gather}
        \mathcal{V}_c := \kappa^{-1}(c) = \bigl\{(\x,y)\in\mathcal{V}\bigm\vert\kappa(\x,y)=c\bigr\}\,.
    \end{gather}
    The algorithm proceeds by constructing a conformal predictor for every class $c\in\mathcal{C}$ with calibration set given by $\mathcal{V}_c$. For every new instance $\mathbf{x}\in\mathcal{X}$ with $\kappa(\mathbf{x},y)=c$, the critical nonconformity score $a_c^*$ is used to construct a prediction interval. Note that every conformal predictor induces an MCP model in a straightforward fashion, where every taxonomy class uses the same nonconformity measure. However, not all MCP models are induced in this way. It is perfectly valid to use a distinct nonconformity measure (or even a different conformal prediction algorithm) for every taxonomy class. Conformal predictors that are not constructed using this Mondrian approach, with a separate conformal prediction instance for every taxonomy class, will be called \textit{non-Mondrian} in this paper.

    The Mondrian approach benefits from the theoretical guarantees of Theorem~\ref{marginal_validity} of the (I)CP algorithm in that validity will be guaranteed for every class in $\mathcal{C}$ individually, as long as the data is exchangeable in every class~\cite{algorithmic}.
    
    \begin{theorem}[Conditional validity]\label{conditional_validity}
        Let $\Gamma^\alpha:\mathcal{X}\times(\mathcal{X}\times\mathbb{R})^\infty\rightarrow[\mathbb{R}]$ be a Mondrian inductive conformal predictor at significance level $\alpha\in[0,1]$ for the taxonomy function $\kappa:\mathcal{X}\times\mathbb{R}\rightarrow\mathcal{C}$. If the nonconformity scores are exchangeable 
        for any calibration set $\mathcal{V}$ and any new observation $(\mathbf{x},y)$, i.e.~any ordering of $A(\mathcal{V})\cup\{A(\mathbf{x},y)\}$ is equally probable, then $\Gamma^\alpha$ is conservatively conditional valid w.r.t.~to the taxonomy function~$\kappa$:
        \begin{gather}
            \mathrm{Prob}\bigl(Y\in\Gamma^\alpha(X,V)\mid\kappa(X,Y)=c\bigr)\geq1-\alpha\,,
        \end{gather}
        for all $c\in\mathcal{C}$, where the probability is taken over both $(X,Y)$ and $V$.
    \end{theorem}

%% file: Theory.tex
\section{Theoretical results}\label{section:uncertainty}

    Although Mondrian conformal prediction has strong conditional guarantees, the calibration set has to be split, something that might become problematic in settings with limited data or a large number of taxonomy classes. In this section, it will be investigated when Mondrian conformal prediction and, hence, a data split, can be avoided by studying the conditional validity of non-Mondrian conformal predictors. In particular, the conditional validity of non-Mondrian conformal predictors is studied from a theoretical perspective. First, the general case is considered, where no parametric assumptions are made about the nonconformity measures, and is related to Fisher's notion of pivotal statistics. Then, by making some stricter assertions, an explicit expression, Eq.~\eqref{general_form}, is derived for the data-generating process such that conditional validity holds w.r.t.~any uncertainty-dependent taxonomy function.

\subsection{Pivotal quantities}

    In the situation at hand, the hope for non-Mondrian conformal prediction is that the heteroskedasticity in a data set can be treated by choosing a suitable nonconformity measure, such as the normalized one~\eqref{normalized_score}, without having to resort to conditional methods such as MCP, since these are more data intensive (and data sparsity is often higher in regions with high heteroskedastic noise). However, there is an important theoretical barrier that limits the usefulness of this pursuit. Namely, if a non-Mondrian model should be able to produce conditionally valid intervals with respect to some taxonomy function, the critical nonconformity scores should be distributed equally across all classes of the chosen taxonomy. The following theorem presents a sufficient condition for conditionally valid conformal predictors. The joint distribution over calibration sets and test instances, which is assumed to exchangeable, will be denoted by $P$.

    \begin{theorem}[Independence]\label{theorem:independence}
        Consider a taxonomy function $\kappa:\mathcal{X}\times\mathbb{R}\rightarrow\mathcal{C}$. If both the distribution of the nonconformity scores and the distribution of the calibration sets are independent of the taxonomy class of the test instance, i.e.
        \begin{gather}
            P_{A\mid C}(B\mid c)=P_A(B)\qquad\mbox{and}\qquad P(\mathcal{V}\mid c)=P(\mathcal{V})
        \end{gather}
        for all $c\in\mathcal{C}$, $B\subseteq\mathbb{R}$ and $\mathcal{V}\in(\mathcal{X}\times\mathbb{R})^n$, the conformal predictor associated with $A$ is conditionally valid w.r.t.~$\kappa$ (Definition~\ref{validity}).
        \begin{proof}
            For any taxonomy class $c\in\mathcal{C}$, the following relation holds:
            \begin{gather*}
                \mathrm{Prob}\bigl(Y\in\Gamma^\alpha(X,V)\,\bigm\vert\,C=c\bigr)=\int_{(\mathcal{X}\times\mathbb{R})^n}\mathrm{Prob}\bigl(Y\in\Gamma^\alpha(X,\mathcal{V})\,\bigm\vert\,C=c,V=\mathcal{V}\bigr)\,\dr P(\mathcal{V}\mid C=c)\,.
            \end{gather*}
            By the definition of ICPs, the first factor can be expressed in terms of the distribution of nonconformity scores:
            \begin{gather*}
                \mathrm{Prob}\bigl(Y\in\Gamma^\alpha(X,V)\,\bigm\vert\, C=c\bigr)=\int_{(\mathcal{X}\times\mathbb{R})^n}F_{A\mid C}\bigl(a^*_{\mathcal{V}}\,\bigm\vert\,C=c,V=\mathcal{V}\bigr)\,\dr P(\mathcal{V}\mid C=c)\,,
            \end{gather*}
            where $a^*_{\mathcal{V}}:=q_{(1-\alpha)\left(1 + \frac{1}{n}\right)}\bigl(A(\mathcal{V})\bigr)$. By assumption, neither the distribution of nonconformity scores nor the distribution of calibration sets depends on the taxonomy class $\kappa(X,Y)$, hence
            \begin{align*}
                \mathrm{Prob}\bigl(Y\in\Gamma^\alpha(X,V)\,\bigm\vert\,C=c\bigr)&=\int_{(\mathcal{X}\times\mathbb{R})^n}F_A\bigl(a^*_{\mathcal{V}}\bigm\vert V=\mathcal{V}\bigr)\,\dr P(\mathcal{V})\\
                &=\mathrm{Prob}\bigl(Y\in\Gamma^\alpha(X,V)\bigr) \geq 1-\alpha\,.
            \end{align*}
            This concludes the proof.
        \end{proof}
    \end{theorem}
    
    \begin{remark}
        Note that the above theorem implies that mere independence of the nonconformity score and taxonomy class of every data point is not sufficient in general. Exchangeability is too weak for the conclusion to hold. In the case of i.i.d.~data, the second condition is, however, trivially satisfied.

        Given that the second condition holds, the first condition is also a very reasonable conditions to require. By exchangeability, the dependence between any two data points is exactly the same. So, if the only the second condition holds, the data process would be such that, for every index $i\leq n$, the nonconformity scores of all data points $(X_j,Y_j)$ except for $i=j$ depend on the taxonomy class of $(X_i,Y_i)$.
    \end{remark}

    When conditional validity is not just required w.r.t.~a fixed taxonomy function, but w.r.t.~an entire family of taxonomy functions, a concept from the classical statistical literature becomes relevant~\cite{arnold1984pivotal,toulis2017useful}.
    
    \begin{definition}[Pivotal quantity]\label{pivotal_quantity}
        Consider a family of distributions $\{P_\theta\mid\theta\in\Theta\}$, parametrized by a set $\Theta$. A function $g$ of observations is called a \textit{pivotal quantity} (or simply \textit{pivot}) if its distribution does not depend on the particular choice of parameter:
        \begin{gather}
            (\forall\theta,\theta'\in\Theta) ((X\sim P_\theta \land X'\sim P_{\theta'}) \Rightarrow g(X)\overset{d}{=}g(X'))\,,
        \end{gather}
        where $\overset{d}{=}$ denotes equality in distribution, i.e.~$g(X)$ and $g(X')$ have the same distribution. The distribution of such a pivotal quantity will be called the \textit{pivotal distribution} further on.
    \end{definition}
    
    Combined with the Independence Theorem~\ref{theorem:independence}, this gives rise to the following result, which states how assuming the conditional validity of a conformal predictor is tied to making assumptions about the data-generating distribution.
    \begin{result}\label{theorem:pivotal}
        A conformal predictor associated with a nonconformity measure $A:\mathcal{X}\times\mathcal{Y}\rightarrow\mathbb{R}$ is conditionally valid for any \textit{feature-dependent Mondrian taxonomy}, i.e.~a Mondrian taxonomy that only depends on the feature space $\mathcal{X}$, if $A$ is a pivotal quantity for the family of conditional distributions $\bigl\{P_{Y\mid X}(\cdot\mid\x)\,\big\vert\,\x\in\mathcal{X}\bigr\}$ and if the data is such that the calibration set $\mathcal{V}$ does not depend on the taxonomy class of any test point, i.e.~$P(\mathcal{V}\mid c)=P(\mathcal{V})$ for all $c\in\mathcal{C}$ and $\mathcal{V}\in(\mathcal{X}\times\mathbb{R})^n$.
    \end{result}

\subsection{Normalization}

    Although the theorems in the preceding section do not make any assumptions about the form of the conditional distribution $P_{Y\mid X}$, beyond the nonconformity measure $A:\mathcal{X}\times\mathcal{Y}\rightarrow\mathbb{R}$ being pivotal, it is interesting to look at some more specific examples. In this section, it will be shown that a large class of common distributions gives rise to often-used nonconformity measures.

    Recall the mean-variance estimators in Example~\ref{mve} with normalized nonconformity measure~\eqref{ncp_score}. If the absolute value was not present in that equation, then the nonconformity scores would actually be the (estimated) $z$-scores of the data. Assuming the idealized situation where the \textit{oracle} can be accessed, i.e.~both the (conditional) mean and variance can be modelled perfectly, such a transformation would lead to the random variable~$A$, the nonconformity score, following a distribution with
    \begin{gather}
        \mathrm{E}[A] = 0\qquad\text{and}\qquad\mathrm{Var}[A] = 1
    \end{gather}
    independent of whether one considers the marginal distribution or conditions on a specific taxonomy class. Using the terminology of Definition~\ref{pivotal_quantity}, this can be rephrased by saying that $A$ is a pivotal quantity. The next theorem shows that such a situation holds more generally. In the remainder of the text, $f_A$ and $f_{A\mid C}$ denote the probability density functions of the distributions $P_A$ and $P_{A\mid C}$, respectively.
    
    \begin{theorem}[Standardization]\label{theorem:standardization}
        If the conditional distribution $P_{Y\mid X}$ has a density function of the form
        \begin{gather}
            \label{general_form}
            f_{Y\mid X}(y\mid\mathbf{x}) = \frac{1}{\sigma(\mathbf{x})}g\left(\frac{y-\mu(\mathbf{x})}{\sigma(\mathbf{x})}\right)
        \end{gather}
        for some smooth function $g:\mathbb{R}\rightarrow\mathbb{R}^+$, then the probability distribution of the standardized nonconformity measure
        \begin{gather}
            \label{standardization}
            A_\text{\upshape st}(\mathbf{x},y) := \frac{y-\mu(\mathbf{x})}{\sigma(\mathbf{x})}
        \end{gather}
        is independent of the classes of any feature-dependent Mondrian taxonomy $\kappa:\mathcal{X}\times\mathcal{Y}\rightarrow\mathcal{C}$.
        \begin{proof}
            The joint density of nonconformity scores and taxonomy classes can be rewritten as
            \begin{align*}
                f_{A_\text{st},C}(a,c) 
                &= \frac{\partial}{\partial a} F_{A_\text{st},C}(a,c)\\
                &= \frac{\partial}{\partial a} 
                   P_{X,Y}\bigl(\{(\mathbf{x},y) \mid A_\text{st}(\mathbf{x},y)\in\,]-\!\infty,a]\land\kappa(\mathbf{x})=c\}\bigr)\\
                &=\frac{\partial}{\partial a}\int_{\kappa^{-1}(c)}\int_{-\infty}^{\mu(\x)+a\sigma(\x)}f_{X,Y}(\x,y)\,\dr y\,\dr\x\\
                &=\int_{\kappa^{-1}(c)}f_{X,Y}\bigl(\x,\mu(\x)+a\sigma(\x)\bigr)\sigma(\x)\,\dr\x\,,
            \end{align*}
            where in the last step Leibniz's integral rule was applied:
            \begin{gather*}
                \frac{\dr}{\dr x}\int_{l(x)}^{u(x)}f(x,y)\,\dr y = f\bigl(x,u(x)\bigr)\frac{\dr u}{\dr x}-f\bigl(x,l(x)\bigr)\frac{\dr l}{\dr x} + \int_{l(x)}^{u(x)}\frac{\partial f}{\partial x}(x,y)\,\dr y\,.
            \end{gather*}
            Finally, the joint density function $f_{X,Y}$ is factorized as
            \begin{gather*}
                f_{X,Y}(\mathbf{x},y) = f_{Y\mid X}(y\mid\mathbf{x})f_X(\mathbf{x})
            \end{gather*}
            to obtain:
            \begin{gather*}
                f_{A_\text{st},C}(a,c)=\int_{\kappa^{-1}(c)}\underbrace{\sigma(\mathbf{x})f_{Y\mid X}\bigl(\mu(\mathbf{x})+a\sigma(\mathbf{x})\,\big\vert\,\mathbf{x}\bigr)}_{=f_{A_\text{st}\mid X}(a\mid\mathbf{x})}f_X(\mathbf{x})\,\dr\mathbf{x}.
            \end{gather*}
            If the first factor is (functionally) independent of $\mathbf{x}$, hence, of $\mu$ and $\sigma$, it can be moved out of the integral:
            \begin{gather*}
                f_{A_\text{st},C}(a,c)=f_{A_\text{st}}(a)\int_{\kappa^{-1}(c)}f_X(\mathbf{x})\,\dr\mathbf{x}=f_{A_\text{st}}(a)P_C(c).
            \end{gather*}
            To see when this holds, define a three-parameter function $\widetilde{f}$ as a generalization of $f_{Y\mid X}$ as follows: \[\widetilde{f}\big(a,\mu(\mathbf{x}),\sigma(\mathbf{x})\big):=f_{Y\mid X}\bigl(\mu(\mathbf{x})+a\sigma(\mathbf{x})\,\big\vert\,\mathbf{x}\bigr)\,.\] If after standardization the density should only depend on $y$, the following system of partial differential equations should hold:
            \begin{gather*}
                \begin{cases}
                    \partial_\mu\bigl(\sigma\widetilde{f}(a,\mu,\sigma)\bigr)=0\,,\\[.2cm]
                    \partial_\sigma\bigl(\sigma\widetilde{f}(a,\mu,\sigma)\bigr)=0\,.
                \end{cases}
            \end{gather*}
            The first partial differential equation immediately yields that $\widetilde{f}$ is independent of $\mu$. Analogously, the second equation says that $\sigma\widetilde{f}$ is independent of $\sigma$ or, equivalently, that
            \begin{gather*}
                \widetilde{f}(a,\mu,\sigma) = \frac{g(a)}{\sigma}
            \end{gather*}
            for an arbitrary function $g:\mathbb{R}\rightarrow\mathbb{R}^+$ (requiring that $\widetilde{f}$ gives a density function imposes further conditions on $g$). Transforming back to the original function gives
            \begin{gather*}
                f_{Y\mid X}(y\mid\mathbf{x}) = \frac{1}{\sigma(\mathbf{x})}g\left(\frac{y-\mu(\mathbf{x})}{\sigma(\mathbf{x})}\right)\,,
            \end{gather*}
            which concludes the proof.
        \end{proof}
    \end{theorem}

    In view of Corollary~\ref{theorem:pivotal}, the standardized variable $A_\text{st}(\x,y)$ is a pivotal quantity for variables coming from the $\mathcal{X}$-parametrized family of distributions~\eqref{general_form} and the function $g:\mathbb{R}\rightarrow\mathbb{R}^+$ is exactly its pivotal distribution.

    \begin{figure}[t]
        \centering
        \includegraphics[width = .9\linewidth]{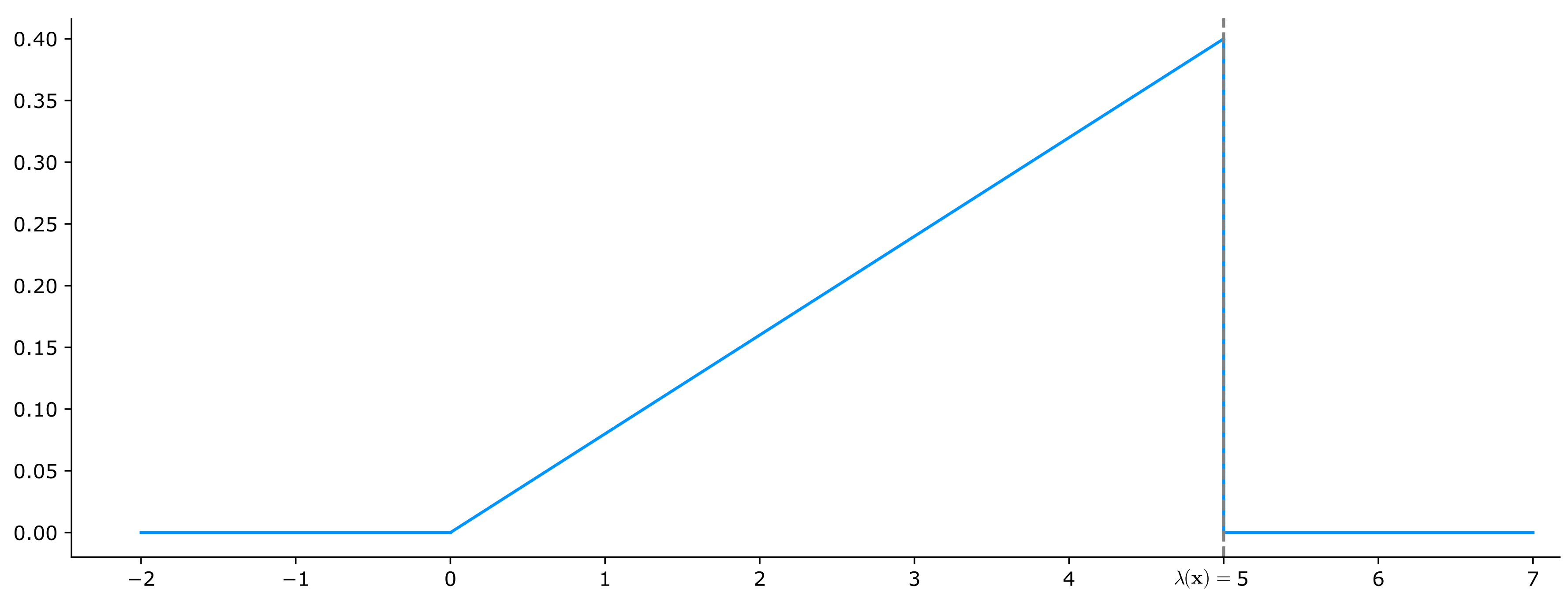}
        \caption{Probability density function of the triangular distribution~\eqref{triangle} with (conditional) width parameter $\lambda(\mathbf{x})=5$.}
        \label{fig:triangle}
    \end{figure}
    
    Although the standardized measure~\eqref{standardization} is not exactly the same as the normalized measure~\eqref{ncp_score}, it is also of interest on its own. It is for example used in the construction of \textit{conformal predictive systems}~\cite{vovk2017nonparametric}, where instead of calibrating at a single quantile, the whole predictive distribution is modelled. The results in this paper carry over to that setting accordingly. Notwithstanding that requiring invariance under standardization seems weaker than what is required to handle the normalized nonconformity measure~\eqref{ncp_score}, the following remark shows that this difference is actually irrelevant.
    
    \begin{remark}\label{theorem:transformation}
        Distributions with a density function of the form 
        \eqref{general_form} also lead to a pivotal distribution for the $\sigma$-normalized residual measure~\eqref{ncp_score}. More generally, if a nonconformity measure $A$ is pivotal, any nonconformity measure obtained by (post)composing it with a feature-independent function will also be pivotal.
    \end{remark}

    \begin{remark}
        Note that, instead of making a detour via the preceding theorem, the proof of Theorem~\ref{theorem:standardization} could have been generalized to work directly with the normalized residual measure~\eqref{ncp_score}: in the step before applying Leibniz's integral rule, the lower integration bound for $y$ given by $-\infty$, would then become $\mu(\x)-a\sigma(\x)$ for the normalized nonconformity measure. The integral rule would then lead to an additional term $\sigma(\x)f\big(\x,\mu(\x)-a\sigma(\x)\big)$. However, Remark~\ref{theorem:transformation} is more generally applicable for any nonconformity measure.
    \end{remark}

    \begin{example}
        Some common examples of distributions, where $\mu$ and $\sigma$ represent the conditional mean and standard deviation, with the following pivotal distributions satisfy the above theorem:
        \begin{enumerate}
            \itemindent=10pt
            \item[(i)] Normal distribution:
                \begin{gather}
                    g(x) = \frac{1}{\sqrt{2\pi}}\exp(-x^2/2)\,.
                \end{gather}
            \item[(ii)] Laplace distribution:
                \begin{gather}
                    g(x) = \frac{1}{\sqrt{2}}\exp(-\sqrt{2}|x|)\,.
                \end{gather}
            \item[(iii)] Uniform distribution:
                \begin{gather}
                    g(x) = \frac{1}{2\sqrt{3}}\mathbbm{1}_{[-\sqrt{3},\sqrt{3}]}(x)\,,
                \end{gather}
                where the indicator function $\mathbbm{1}$ was defined in Eq.~\eqref{indicator_function}.
        \end{enumerate}
        To give an example of how general the allowed distributions in this theorem are, consider the following asymmetric one (see Fig.~\ref{fig:triangle}):
        \begin{gather}
            \label{triangle}
            f_{Y\mid X}(y\mid\mathbf{x})=\frac{2y}{\lambda(\mathbf{x})^2}\mathbbm{1}_{[0,\lambda(\mathbf{x})]}(y)
        \end{gather}
        for some positive function $\lambda:\mathcal{X}\rightarrow\mathbb{R}^+$. Although it is seemingly not of the form~\eqref{general_form}, it can, with some work, be rewritten as such:
        \begin{gather}
            \frac{2y}{\lambda(\mathbf{x})^2}\mathbbm{1}_{[0,\lambda(\mathbf{x})]}(y) = \frac{1}{\sigma(\mathbf{x})}\left(\frac{1}{9}\left(\frac{y-\mu(\mathbf{x})}{\sigma(\mathbf{x})}\right)+\frac{2\sqrt{2}}{9}\right)\mathbbm{1}_{[-2\sqrt{2},\sqrt{2}]}\left(\frac{y-\mu(\mathbf{x})}{\sigma(\mathbf{x})}\right)\,,
        \end{gather}
        where
        \begin{gather}
            \mu(\mathbf{x}) := \frac{2\lambda(\mathbf{x})}{3} \qquad\mbox{and}\qquad \sigma(\mathbf{x}) := \frac{\lambda(\mathbf{x})}{3\sqrt{2}}\,.
        \end{gather}
    \end{example}

    Essentially, to obtain a pivotal distribution for the standardized (or normalized) nonconformity measure, the conditional distribution $P_{Y\mid X}$ should be obtained as a member of the location-scale family with parameters $\mu(\mathbf{x})$ and $\sigma(\mathbf{x})$ induced by the distribution $g$. As a consequence, requiring invariance under standardization also allows for conditional distributions such as exponential distributions. Exponential distributions in general only form a scale family and not a location family. However, because $\mu=\sigma$ for exponential distributions, the conditional version does form a location-scale family generated by $\mu$ and~$\sigma$. Accordingly, it gives rise to the following pivotal distribution for standardized variables:
    \begin{gather}
        g(x) = \exp(-x-1)\theta(x+1)\,,
    \end{gather}
    where $\theta:\mathbb{R}\rightarrow\{0,1\}$ denotes the Heaviside step function:
    \begin{gather}
        \theta(x) := \mathbbm{1}_{[0,+\infty[}(x)\,.
    \end{gather}

    Note that the functions $\mu:\mathcal{X}\rightarrow\mathbb{R}$ and $\sigma:\mathcal{X}\rightarrow\mathbb{R}^+$ in the preceding theorems in general do not have to be the conditional mean and standard deviation. Moreover, it is not even strictly necessary that the parameters $\mu$ and $\sigma$ are estimated perfectly. For example, from the form of Eq.~\eqref{general_form} it can immediately be seen that the estimators $\widehat{\mu}$ and $\widehat{\sigma}$ only need to satisfy
    \begin{align}
        \label{std_estimation}
        \widehat{\sigma} &= \lambda\sigma\,,\\
        \label{mean_estimation}
        \widehat{\mu}-\mu &= \lambda'\widehat{\sigma}\,,
    \end{align}
    for some $\lambda\in\mathbb{R}^+$ and $\lambda'\in\mathbb{R}$. (These relaxations also follow from Remark~\ref{theorem:transformation}.)

    \begin{example}[Additive noise]
        The above theorem also shows that any data set having a conditional generating process of the form~\cite{james2013introduction,johansson2014regression,lei}
        \begin{gather}
            y = \mu(\mathbf{x}) + \sigma(\mathbf{x})\varepsilon\,,
        \end{gather}
        where $\varepsilon$ is sampled from some fixed distribution, will lead to a conditionally valid NCP algorithm. In other words, NCP models will be conditionally valid for any data set obtained by adding noise to a fixed trend. It is important to remark that the distribution of $\varepsilon$ is unconstrained and does not have to be of the above functional form.
    \end{example}

    To round off this section, it can be useful for future work to note that the method of proof of Theorem~\ref{theorem:standardization} can be generalized to other nonconformity measures.
    
    \begin{theorem}
        Assume that $P_{Y\mid X}$ admits a density function $f_{X\mid Y}$, smoothly parameterized by the parameters $\boldsymbol{\theta}\equiv\boldsymbol{\theta}(\x)$, and assume that the nonconformity measure $A:\mathcal{X}\times\mathbb{R}\rightarrow\mathbb{R}$ only depends on $\mathcal{X}$ through $\boldsymbol{\theta}$. If, for a fixed $\x\in\mathcal{X}$, the transformation $y\mapsto A(\x,y)$ is increasing, then $A$ is pivotal if the following condition is satisfied:
        \begin{gather}
            f_{Y\mid X}\bigl(g(a,\boldsymbol{\theta});\boldsymbol{\theta}\mid\x\bigr)\nabla_{\boldsymbol{\theta}}\frac{\partial g(a,\boldsymbol{\theta})}{\partial a} + \frac{\partial g(a,\boldsymbol{\theta})}{\partial a}\nabla_\theta f_{Y\mid X}\bigl(g(a,\boldsymbol{\theta});\boldsymbol{\theta}\mid\x\bigr) = \mathbf{0}\,,
        \end{gather}
        where $g\bigl(\cdot,\boldsymbol{\theta}\bigr):\mathbb{R}\rightarrow\mathbb{R}$ is defined by the equation
        \begin{gather}
            A\Bigl(g\bigl(a,\boldsymbol{\theta}\bigr),\x\Bigr)=a\,,
        \end{gather}
        i.e.~it is the inverse of $A$ for fixed $\x\in\mathcal{X}$.
    \end{theorem}
    
    This differential equation could be solved, for example, in the case of the interval nonconformity measure in~\eqref{interval_score}. However, this would be less sensible. The normalized nonconformity measure does not take into account the significance level at which the prediction intervals are going to be constructed. It simply gives a statistic of the predictive distribution. However, any interval predictor does assume a predetermined significance level in some way and, hence, it should only become a pivotal quantity at that given significance level.

%% file: Synthetic.tex
\section{Experiments on synthetic data}\label{section:synthetic}

    By performing and analyzing some synthetic experiments, both the results from Section~\ref{section:uncertainty} -- the nonconformity measure being a pivotal quantity is a sufficient condition for conditional validity -- can be validated, and a diagnostic tool can be developed to help assess whether a non-Mondrian conformal predictor could be conditionally valid (w.r.t.~a given taxonomy function).

\subsection{Data types}

    To compare the different methods in a controlled manner, some synthetic data sets are considered. Four different types are of importance and representative of real-world situations:
    \begin{enumerate}[\qquad\qquad]
        \item[Type 1.] constant mean: $\mathrm{E}[Y\mid X]=\mathrm{E}[Y]$,
        \item[Type 2.] functional dependence: $\mathrm{Var}[Y\mid X]=\varphi\bigl(\mathrm{E}[Y\mid X]\bigr)$ for some function $\varphi:\mathbb{R}\rightarrow\mathbb{R}^+$,
        \item[Type 3.] low-dimensional representation: $\mathrm{Var}[Y\mid X]=f(X^\downarrow)$, where $X^\downarrow$ denotes the projection of $X$ onto a subspace of $\mathcal{X}$, such as the projection onto the first component, and
        \item[Type 4.] mixture models.
    \end{enumerate}
    Note that, similar to Type 1 data, one could also consider data sets with constant variance. However, this implies homoskedasticity (at least aleatorically) and, hence, is not considered here.
    
    To generate these data sets synthetically, the main sampling procedure is as follows:
    \begin{enumerate}[\qquad]
        \item A parametric family of distributions $P_{Y\mid X}(\,\cdot\,;\mu,\sigma^2\mid\cdot)$ is fixed.
        \item $n\in\mathbb{N}$ feature tuples $\x$ are sampled from a fixed distribution $P_X$, e.g.~a uniform distribution over a $k$-dimensional (unit) hypercube.
        \item Mean and variance functions $\mu:\mathcal{X}\rightarrow\mathbb{R}$ and $\sigma^2:\mathcal{X}\rightarrow\mathbb{R}^+$ are fixed.
        \item For every feature tuple $\x\in\mathcal{X}$, a response $y$ is sampled from the distribution $P_{Y\mid X}\bigl(
        \cdot\mid \x;\mu(\x),\sigma^2(\x)\bigr)$.
    \end{enumerate}

    To evaluate the quality of interval predictors $\Gamma^\alpha:\mathcal{X}\rightarrow[\mathbb{R}]$, two performance metrics are used: the \textit{empirical coverage} and the \textit{average size of the prediction regions}~\cite{dewolf2023valid}. Given a joint distribution $P_{X,Y}$ on $\mathcal{X}\times\mathbb{R}$, the coverage is defined as follows:
    \begin{gather}
        \label{coverage}
        \mathcal{C}(\Gamma^\alpha,P_{X,Y}) := \mathrm{E}\left[\mathbbm{1}_{\Gamma^\alpha(X)}(Y)\right] = \mathrm{Prob}\bigl(Y\in\Gamma^\alpha(X)\bigr)\,,
    \end{gather}
    thereby turning Definition~\ref{validity} into the condition $C(\Gamma^\alpha,P_{X,Y})\geq1-\alpha$. The average width is defined as
    \begin{gather}
        \label{expected_size}
        \mathcal{W}(\Gamma^\alpha,P_{X,Y}) := \mathrm{E}\bigl[|y_+(X) - y_-(X)|\bigr]\,,
    \end{gather}
    where the functions $y_\pm:\mathcal{X}\rightarrow\mathbb{R}$ denote the upper and lower bounds of the prediction intervals produced by $\Gamma^\alpha$. When $P_{X,Y}$ is the empirical distribution of a data set $\mathcal{D}$, the notation $\mathcal{C}(\Gamma^\alpha,\mathcal{D})$ is also used. Of course, since the focus lies on conditional performance, conditional counterparts can be defined as well:
    \begin{gather}
        \label{c_coverage}
        \mathcal{C}(\Gamma^\alpha,P_{X,Y}\mid c) := \mathrm{Prob}\bigl(Y\in\Gamma^\alpha(X)\,\big\vert\,\kappa(X,Y)=c\bigr)
    \end{gather}
    and
    \begin{gather}
        \label{c_expected_size}
        \mathcal{W}(\Gamma^\alpha,P_{X,Y}\mid c) := \mathrm{E}\bigl[|y_+(X) - y_-(X)|\,\big\vert\,\kappa(X,Y)=c\bigr]\,.
    \end{gather}
    Note that whereas the (marginal) distribution over $\mathbb{R}$ is irrelevant for the marginal measures, it plays a role in the conditional definitions, since the taxonomy function can in general also depend on the response variable.

\subsection{Deviations from oracle}

    Up to some very specific relaxations in the form of Eqs.~\eqref{std_estimation} and~\eqref{mean_estimation} and, more generally, Remark~\ref{theorem:transformation}, the theorems and methods from the previous section require the parameters to be estimated exactly (or at least consistently in the large data setting). However, this assumption is not a very realistic one in practice. Estimating higher conditional moments, such as the variance, to high precision usually requires state-of-the-art methods and, even more so, a large amount of data, since without strong parametric assumptions multiple samples with nearly identical features are necessary. This requirement is hard to achieve, especially in high-dimensional settings.

    For this reason it is interesting to see what happens when the estimates deviate from the oracle. In general, two possibilities exist:
    \begin{itemize}[10pt]
        \item \textbf{Misspecification}: In general, the nonconformity measure depends on the estimated parameters. Therefore, if these estimators are misspecified, then the transformation $(X,Y)\mapsto A$ might not remove all dependency on $\mathcal{X}$.
        \item \textbf{Contamination}: As for the nonconformity measure, the taxonomy function will, in general, depend on the estimated parameters. If these estimators are misspecified, then the taxonomy classes can get mixed up and even though the distribution of the nonconformity scores might not depend on the true taxonomy, it might depend on the estimated taxonomy. 
    \end{itemize}

    A simple illustration will be enlightening at this point.
    \begin{example}
        Recall Example~\ref{problem_example} and consider the residual nonconformity measure in Eq.~\eqref{residual_score}:
        \begin{gather}
            A(\x,y) = |y-\widehat{\mu}(\x)|\,.
        \end{gather}
        The taxonomy function is still the indicator function in Eq.~\eqref{indicator_function}:
        \begin{gather}
            \kappa_\xi(\x,y) = \mathbbm{1}_{[0,\xi]}(x_2)\,,
        \end{gather}
        where $\xi\in\mathbb{R}$. For $\xi=0.5$, it is not hard to see that the conformal predictor associated with $A$ will be conditionally valid, even though $A$ itself is not pivotal for the given data-generating process.

        Analyzing the effect of misspecification and contamination is now also quite straightforward. If the location $\widehat{\mu}$ is misspecified (in such a way that the residuals have a mean that depends on $x_2$), then the distribution of the nonconformity scores will also depend on the taxonomy, no matter how perfect the parameter $\xi$ is fine-tuned. On the other hand, as soon as $\xi$ deviates from $0.5$, e.g.~when it would be estimated based on a data sample, the distribution of nonconformity scores would also no longer be independent of the classes, even when $\widehat{\mu}$ is modelled perfectly.
    \end{example}
    
    For the purpose of this paper, however, where both the taxonomy function and the nonconformity measure depend explicitly on the same estimate of the data noise, see Eq.~\eqref{taxonomy_function}, misspecification and contamination go hand in hand.

\begin{figure}[t!]
        \centering
        \includegraphics[width = .9\linewidth]{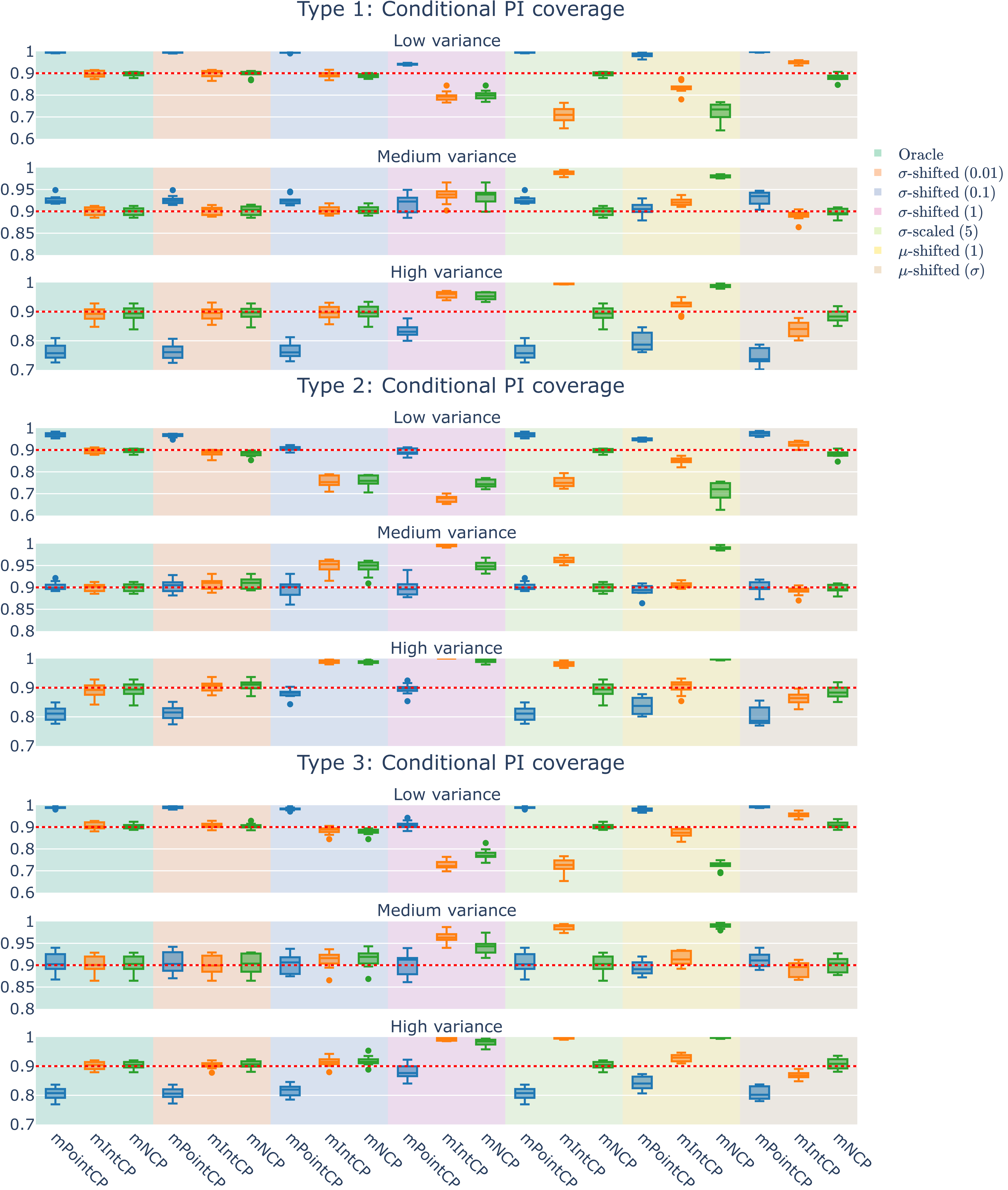}
        \caption{Conditional coverage at significance level $\alpha=0.1$ for synthetic data sets of Types~1, 2 and~3. For every type, the data is divided in three folds based on equal-frequency binning of the estimated variance. The coloured columns indicate the type of misspecification (from left to right): oracle, additive noise on the standard deviation (means of 0.01, 0.1 and 1), scaling by factor 5 of the standard deviation and additive noise on the mean (means of 1 and $\widehat{\sigma}$). For every model, three nonconformity measures are shown (from left to right): residual, interval and $\widehat{\sigma}$-normalized nonconformity measure.}
        \label{fig:synth_types123}
    \end{figure}

    In Fig.~\ref{fig:synth_types123}, the coverage results for three of the four types of synthetic data are shown (Types 1--3) for different kinds of misspecification and for the three types of nonconformity measure introduced in Section~\ref{section:cp}: the residual~\eqref{residual_score}, interval~\eqref{interval_score} and $\sigma$-normalized measures~\eqref{ncp_score}. For each of these figures, the data-generating distribution $P_{Y\mid X}$ has the general form~\eqref{general_form} of Theorem~\ref{theorem:standardization}. The misspecification is simulated by adding random noise to the values of the mean and variance. The first (green) column, indicated by the label `Oracle', uses the true mean and standard deviation.
    
    The columns (orange, blue and pink) indicated by the label `$\sigma$-shifted ($\lambda$)' show the empirical coverage for increasing values of $\sigma$-noise (these estimates are clipped to $\mathbb{R}^+$ to enforce positivity of the standard deviation):
    \begin{gather}
        \widehat{\sigma}(\x) = \sigma(\x) + \varepsilon\qquad\mbox{and}\qquad\varepsilon\sim\mathcal{N}\left(0,\lambda^2\right)\,.
    \end{gather}
    This simulates the behaviour of models that are not able to estimate the variance consistently. It is clear that for larger values of the noise, the (non-Mondrian) conformal predictor using the normalized conformal measure also stops being conditionally valid. The light green column, indicated by the label `$\sigma$-scaled' shows the coverage when the variance is scaled by a fixed value (5 in this case). As expected from Remark~\ref{theorem:transformation} and, in particular, Eq.~\eqref{std_estimation}, this does not change anything in terms of the conditional validity of the normalized (and, trivially, residual) conformal predictors. However, it does break the conditional validity of the conformal predictor using the interval nonconformity measure. This observation is also to be expected. For the mean-variance estimators from Example~\ref{mve} with a normality assumption, the intervals are of the form~\eqref{mve_interval}. It follows that the interval measure~\eqref{interval_score} in this case can be rewritten as follows:
    \begin{align}
        \max\bigl(\widehat{y}_-(\x)-y,y-\widehat{y}_+(\x)\bigr) &= \max\bigl(\widehat{\mu}(\x)-z^\alpha\widehat{\sigma}(\x)-y,y-\bigl(\widehat{\mu}(\x)+z^\alpha\widehat{\sigma}(\x)\bigr)\bigr) \nonumber\\
        &= \max\bigl(\widehat{\mu}(\x)-y-z^\alpha\widehat{\sigma}(\x),y-\widehat{\mu}(\x)-z^\alpha\widehat{\sigma}(\x)\bigr) \nonumber\\
        &= \max\bigl(\widehat{\mu}(\x)-y,y-\widehat{\mu}(\x)\bigr)-z^\alpha\widehat{\sigma}(\x) \nonumber\\
        &= |\widehat{\mu}(\x)-y|-z^\alpha\widehat{\sigma}(\x)\,.
    \end{align}
    From this last line, it is clear that scaling the variance is not equivalent to applying a feature-independent transformation and, accordingly, Remark~\ref{theorem:transformation} is not applicable.

    The last two columns (yellow and brown), indicated by the label `$\mu$-shifted', show the coverage when the mean is shifted by, respectively, a constant and a value proportional to the standard deviation:
    \begin{gather}
        \widehat{\mu}(\x) = \mu(\x) + \varepsilon\qquad\text{with}\qquad\varepsilon\sim\mathcal{N}\left(0,\lambda^2\widehat{\sigma}(\x)^2\right)\,.
    \end{gather}
    It is immediately clear that whereas the constant shift breaks the conditional validity of all three conformal predictors, a shift proportional to the standard deviation does preserve the conditional validity of the normalized model. This is entirely in line with the above results, in particular Eq.~\eqref{mean_estimation}.
    
\begin{figure}[t!]
        \centering
        \includegraphics[width = .9\linewidth]{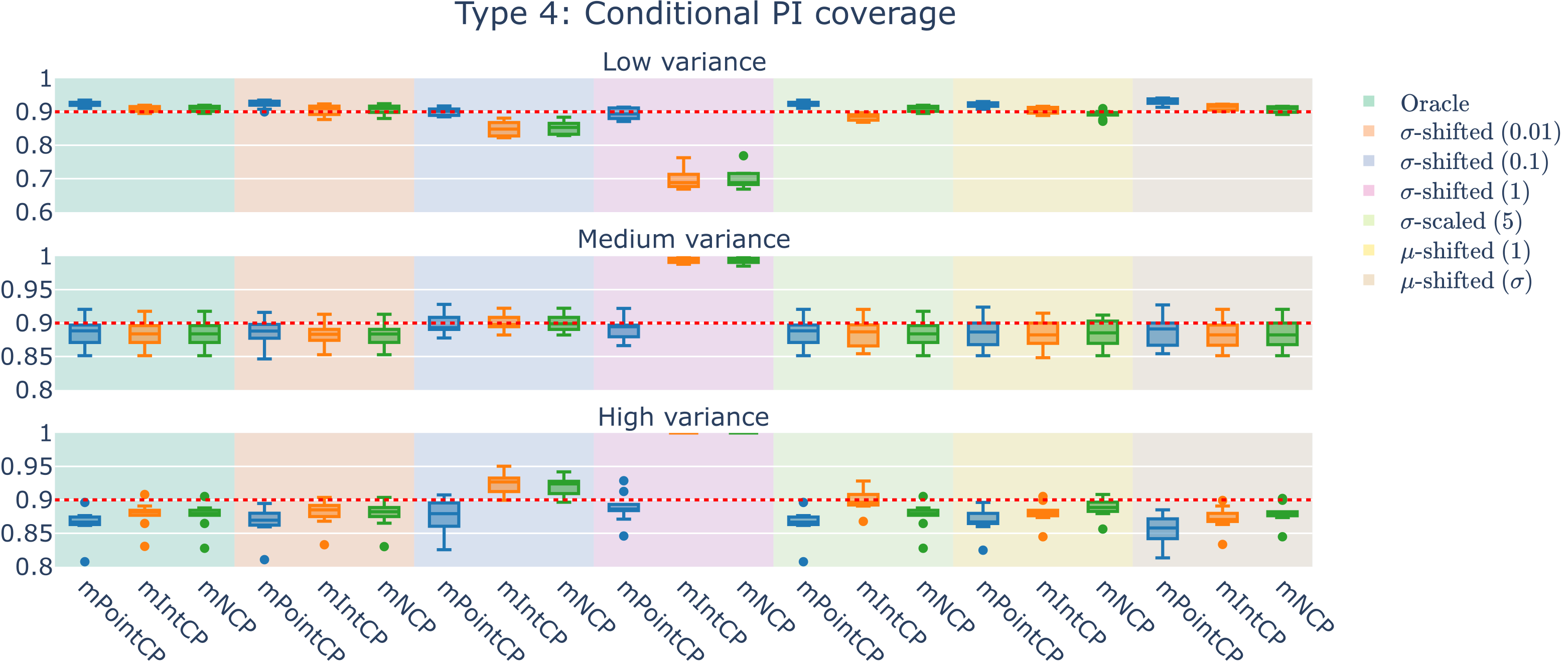}
        \caption{Conditional coverage at significance level $\alpha=0.1$ for a synthetic data set of Type 4. The data is divided in three folds based on equal-frequency binning of the estimated variance. The coloured columns indicate the type of misspecification (from left to right): oracle, additive noise for the standard deviation (means of 0.01, 0.1 and 1), scaling by factor 5 of the standard deviation and additive noise for the mean (means of 1 and $\widehat{\sigma}$). For every model, three nonconformity measures are shown (from left to right): residual, interval and $\widehat{\sigma}$-normalized nonconformity measure.}
        \label{fig:synth_cov_bimodal}
    \end{figure}

    Figure~\ref{fig:synth_cov_bimodal} shows a similar plot for a data set of Type~4. In this case the data is sampled from a bimodal mixture:
    \begin{gather}
        y\sim
        \begin{cases}
            \mathcal{N}\bigl(\mu(\x)-1,0.01\mu(\x)^2\bigr)&\quad\text{if }\mu(\x)\leq2\,,\\[.2cm]
            \mathcal{N}\bigl(\mu(\x)+1,0.01\mu(\x)^2\bigr)&\quad\text{if }\mu(\x)>2\,.
        \end{cases}
    \end{gather}
    Even from the `Oracle'-column it is already clear that the non-Mondrian conformal predictors for such a distribution are not conditionally valid. When moving away from the oracle, the situation is at best sustained. This observation is expected to hold for all mixture distributions $P_{Y\mid X}$. Unless a consistent model can be obtained for every component in the mixture and that each of these components admits the same pivotal distribution, conditional validity will not hold. 

\subsection{Diagnostics}

    The theorem in the preceding section provides a means to get an idea of the conditional behaviour of conformal predictors without actually having to consider a test set. Comparing the distributions of the nonconformity scores over the different strata of the calibration set can give insight into how well the methods will perform and what impact misspecification and contamination might have.

\begin{figure}[t!]
        \centering
        \includegraphics[width = \linewidth, trim = {0 0 0 4cm}, clip]{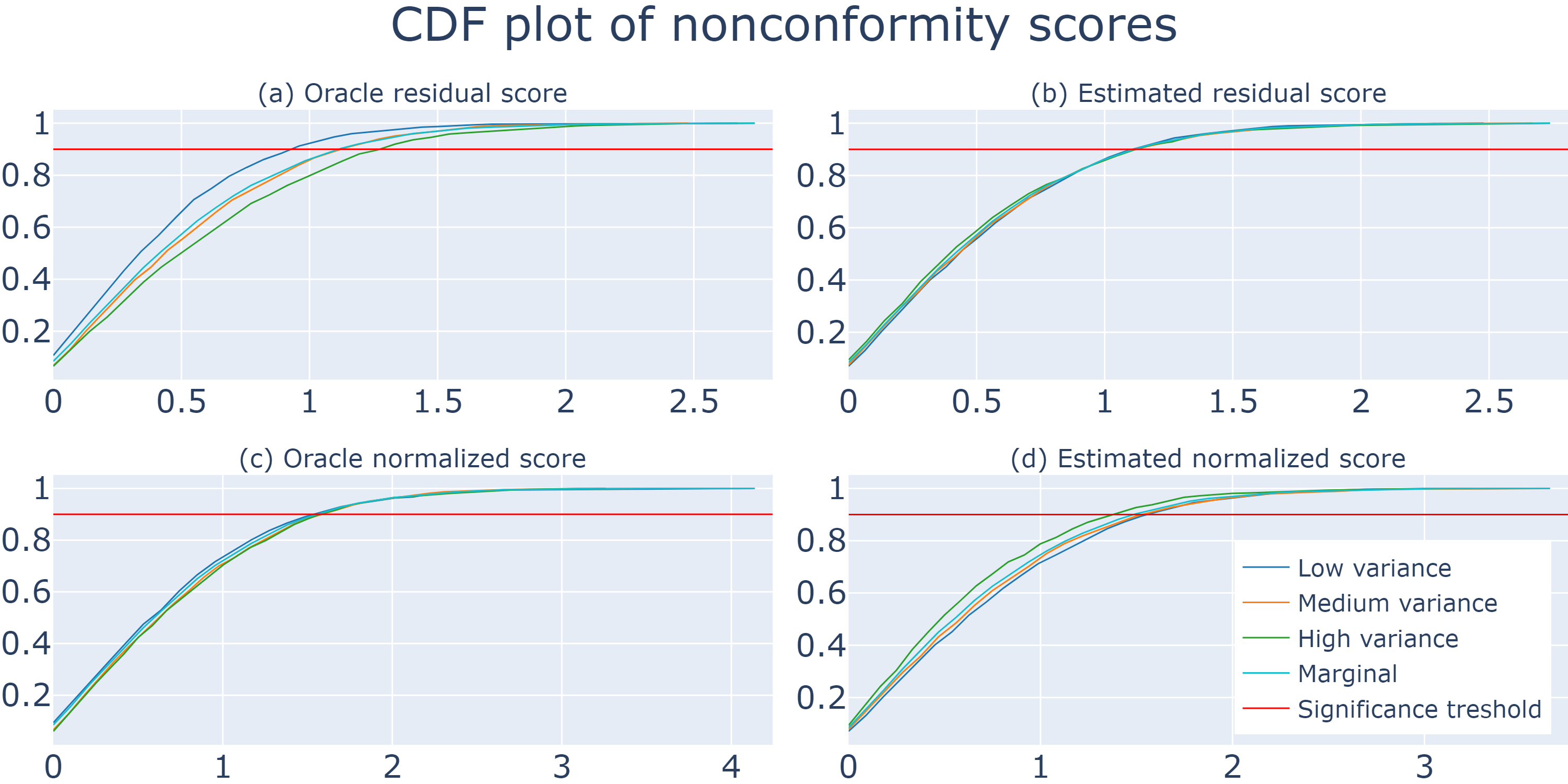}
        \caption{(a) and (c): CDF plots of the residual and $\sigma$-normalized nonconformity scores for the oracle. (b) and (d): CDF plots of the residual and $\sigma$-normalized nonconformity scores for the misspecified model in Equation~\eqref{toy_model} The (empirical) distributions are shown marginally (all data) and for the taxonomy classes corresponding to equal-frequency binning of the estimated variance ($n=3$ classes).}
        \label{fig:cdf_plots}
    \end{figure}

    As a toy example, a family of normal distributions with constant coefficient of variation, fixed at $c_v=0.1$, is chosen as data-generating process:
    \begin{gather}
        \label{toy_model}
        y\sim\mathcal{N}\bigl(\mu(\x),0.01\,\mu(\x)^2\bigr)\,.
    \end{gather}
    Figure~\ref{fig:variance_cdf} shows a CDF plot of the variances of these distributions in case $\mu(\x):=\mathrm{mean}(\x)$ and $\x\sim\mathcal{U}^n(0, 100)$. The colors indicate the taxonomy classes corresponding to equal-frequency binning~\eqref{taxonomy_function} of the variances. In Figs.~\subref{fig:cdf_plots}{a} and~\subref{fig:cdf_plots}{c} the CDF plots of the residual~\eqref{residual_score} and $\sigma$-normalized nonconformity scores~\eqref{ncp_score} are shown, respectively. When applying a non-Mondrian conformal predictor with the residual nonconformity measure to 20 random test sets of 1000 instances, sampled from the above distributions~\eqref{toy_model}, the results shown on the first line of Table~\ref{tab:scores} are obtained.
    
    \begin{subfigures}
        \begin{figure}[t!]
            \centering
            \includegraphics[width = .9\linewidth, trim = {0 0 0 5cm}, clip]{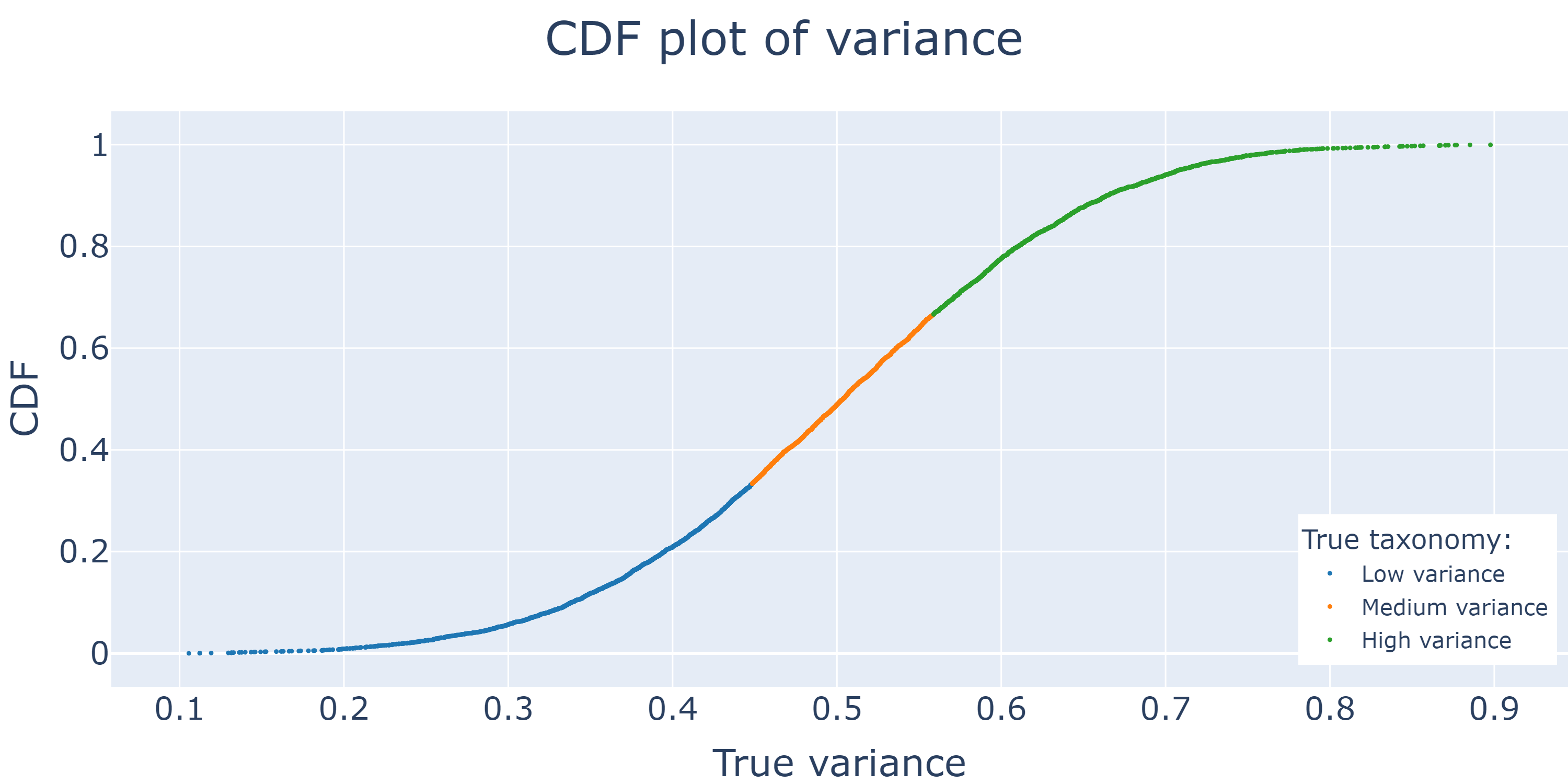}
            \caption{CDF plot of the (true) variance. The colors indicate the taxonomy classes with equal-frequency binning ($n=3$ classes).}
            \label{fig:variance_cdf}
        \end{figure}
        \begin{figure}[t!]
            \centering
            \includegraphics[width = .8\linewidth, trim = {0 0 0 5cm}, clip]{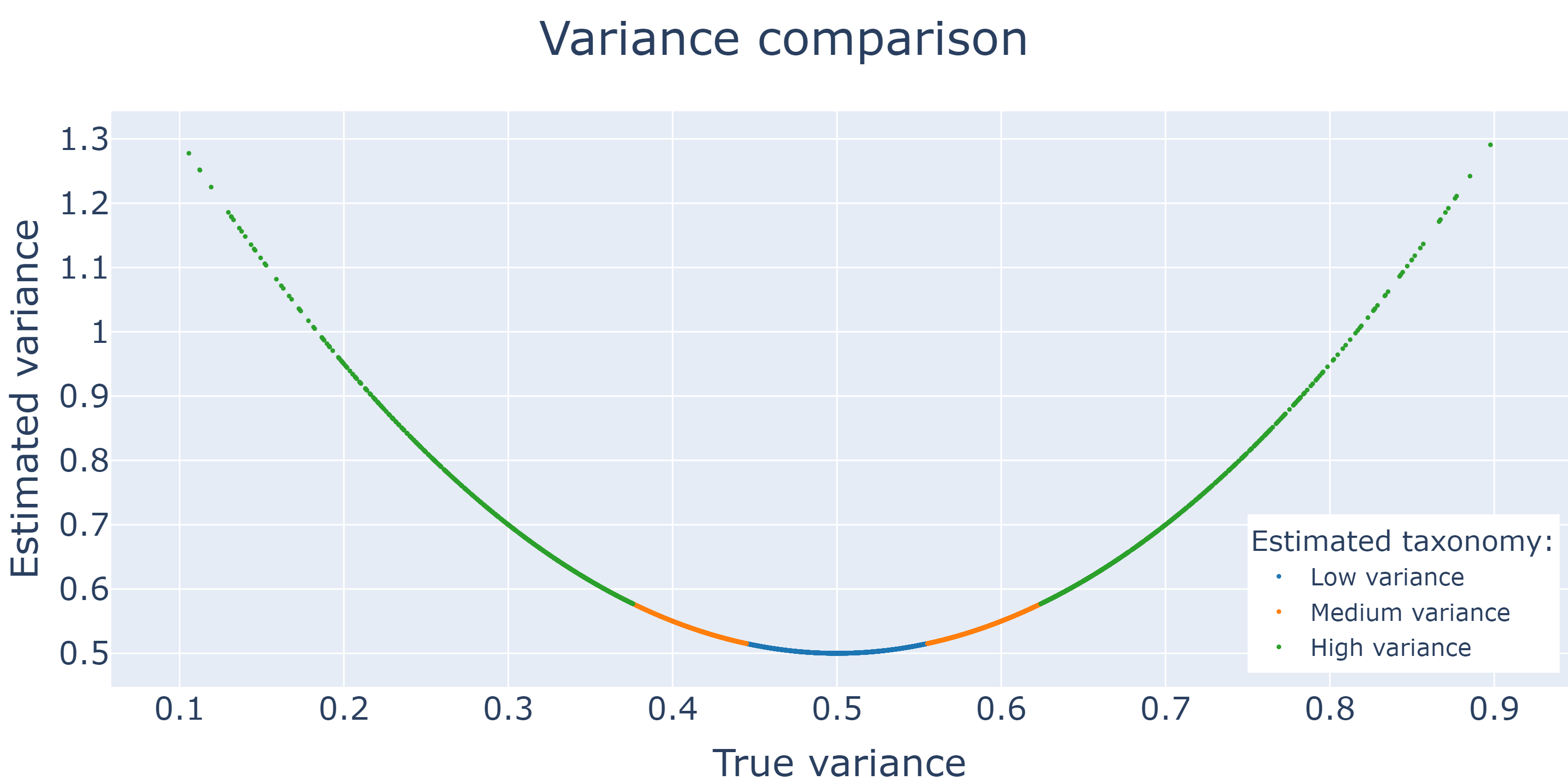}
            \caption{Comparison of the true and estimated variance. The colors indicate the taxonomy classes based on the estimated variance with equal-frequency binning ($n=3$ classes).}
            \label{fig:variance_comparison}
        \end{figure}
    \end{subfigures}

    \begin{table}[t!]
        \caption{Interval coverage degrees for different methods across variance classes (at significance level $\alpha=0.1$). Mean and standard deviation over 20 samples are provided.}
        \label{tab:scores}
        \centering
        \vspace{.5\baselineskip}
        \begin{tabular}{ccccc}
           \toprule
            &Marginal&Low variance&Medium variance&High variance\\
            \midrule
            $A_\text{res}$&$0.905\pm0.008$&$0.951\pm0.010$&$0.904\pm0.017$&$0.856\pm0.018$\\
            $A_\sigma$&$0.902\pm0.008$&$0.905\pm0.018$&$0.902\pm0.015$&$0.898\pm0.017$\\
            \bottomrule
        \end{tabular}
    \end{table}

    While performing experiments on synthetic data, it was observed that most methods showed quadratic deviations when comparing the true variances to the predicted variances. In the region with high levels of noise, the estimates were approximately correct, but in the regions with low noise levels, the estimates were often much larger (of the same magnitude as for high noise). This effect can be modelled by using the following explicitly misspecified model:
    \begin{gather}
        \label{misspecification}
        \widehat{\mu}(\x) := \mu(\x)\qquad\text{and}\qquad\widehat{\sigma}^2(\x) := 5\bigl(\sigma^2(\x) - 0.5\bigr)^2 + 0.5\,.
    \end{gather}
    The consequences of misspecification and contamination can be seen in Fig.~\ref{fig:variance_comparison} for the variance and Fig.~\ref{fig:cdf_plots} for the nonconformity scores. In the first figure, the estimated variance is shown in function of the true variance. The colors again indicate the taxonomy classes, but this time those determined by the estimated variances. As is clearly visible by comparing Figs.~\ref{fig:variance_cdf} and~\ref{fig:variance_comparison}, the taxonomy classes are completely different from what the true classes would be. The true class with medium variance corresponds to the class with low estimated variance and the classes with low and high true variance have been reshuffled to become 50/50 mixtures of low and high estimated variance. When comparing the CDF plots~\subref{fig:cdf_plots}{a} and~\subref{fig:cdf_plots}{c} to~\subref{fig:cdf_plots}{b} and~\subref{fig:cdf_plots}{d}, an interesting effect can be seen. Whereas the residual score for the oracle did not give rise to conditional coverage, it does do so for the misspecified model. On the other hand, for the $\widehat{\sigma}$-normalized nonconformity score, the effect works in the opposite direction. For the oracle, conditional validity is obtained (for all significance levels), but for the misspecified model, only marginal validity is attained. These results are also reflected in Table~\ref{tab:misspecified_scores}.

    \begin{table}[t!]
        \caption{Interval coverage degrees for different methods across variance classes (at significance level $\alpha=0.1$) for the misspecified model~\eqref{misspecification}. Mean and standard deviation over 20 samples are provided.}
        \label{tab:misspecified_scores}
        \centering
        \vspace{.5\baselineskip}
        \begin{tabular}{ccccc}
            \toprule
            &Marginal&Low variance&Medium variance&High variance\\
            \midrule
            $A_\text{res}$&$0.905\pm0.008$&$0.905\pm0.017$&$0.905\pm0.014$&$0.905\pm0.012$\\
            $A_\sigma$&$0.905\pm0.008$&$0.886\pm0.017$&$0.897\pm0.014$&$0.931\pm0.008$\\
            \bottomrule
        \end{tabular}
    \end{table}

    The analysis of the toy model~\eqref{toy_model} leads to the following diagnostic method.
    
    \begin{method}[CDF plots]
        Assuming that the conformal predictors are not overly conservative, meaning that ties do not arise by Theorem~\ref{marginal_validity}, and that the calibration set is representative, creating a CDF plot of both the marginal nonconformity scores and the conditional nonconformity scores for all taxonomy classes can give an idea of how well the marginal method will perform conditionally.
        
        As is immediately clear from Fig.~\subref{fig:cdf_plots}{a}, at significance level $\alpha=0.1$, the quantiles for low and high variance are, respectively, smaller and greater than the marginal quantile. This directly translates to over- and undercoverage in these regions, although marginally the model is valid (as expected from Theorem~\ref{marginal_validity}). Accidentally, the CDF for the taxonomy class with medium variance intersects the marginal CDF at around this significance level and this is also visible in the coverage values. For medium variance, the coverage is close to the nominal level.

        For the normalized conformal predictor with the scores from Fig.~\subref{fig:cdf_plots}{c}, the coverage degrees are shown on the second line of Table~\ref{tab:scores} (again for 20 test sets with 1000 instances). The fact that all CDFs approximately coincide at all levels in the figure is reflected in the conditional coverage values in the table. The model is valid for all classes.
    \end{method}

    Another analytical tool can be used in the situation where a single calibration set is used and, accordingly, consistency of the $(1-\alpha)$-sample quantile is assumed.
    
    \begin{method}[Bootstrap analysis]
        In cases in which visual inspection of the CDF plots does not give a clear interpretation, a statistical test can be used to determine whether the required quantiles coincide for the different taxonomy classes. To compare the $(1-\alpha)$-quantiles between two different taxonomy classes, the method from Wilcox \textit{et al.}~\cite{wilcox2014comparing} can be used. Choose two taxonomy classes $c_1,c_2\in\mathcal{C}$ and consider a fixed number of bootstrap samples $\{\mathcal{V}^1_i\}_{i=1,\ldots,B}$ and $\{\mathcal{V}^2_i\}_{i=1,\ldots,B}$, sampled from the data sets $\mathcal{V}_{c_1}$ and $\mathcal{V}_{c_2}$, respectively. For every $i\in\{1,\ldots,B\}$, the sample difference
        \begin{gather}
            d_i:=\widehat{q}_{(1-\alpha)\left(1 + \frac{1}{n}\right)}(\mathcal{V}^1_i)-\widehat{q}_{(1-\alpha)\left(1 + \frac{1}{n}\right)}(\mathcal{V}^2_i)
        \end{gather}
        can be calculated using the Harrell--Davis quantile estimator~\cite{harrell1982new}. A(n) (approximate) bootstrap confidence interval, at significance level $\beta\in[0,1]$, is given by $[d_{(B\beta/2)},d_{(B-B\beta/2)}]$. If 0 is not contained in the confidence interval for a suitable value of $\beta$, e.g.~$\beta=0.025$, evidence has been found against the use of a marginal conformal predictor for obtaining conditional validity.
    \end{method}

    \begin{remark}
        The above diagnostic tools, especially the bootstrap analysis, is primarily useful in case only a fixed significance level (or a finite number of them) is of interest. To test whether validity will hold at a range of significance levels (or even at all levels), another test for comparing distributions, such as the Kolmogorov--Smirnov test, might be preferred.
    \end{remark}

%% file: Real.tex
\section{Experiments on real data}\label{section:experiment}

    In the synthetic experiments of the previous section, it was possible to investigate the impact of misspecification. In practice, however, there is no way to know the exact model parameters and misspecification almost always occurs. For this reason it is also useful to see how realistic, data-driven models perform.

\subsection{Models}\label{section:models}

    Since the focus lies on uncertainty-dependent conditioning, only models that actually provide estimates of the residual variance will be considered. This excludes point predictors such as standard neural networks or random forests. The methods considered in this benchmarking effort are listed below (abbreviations that will be used in the remainder of the text are indicated in between parentheses):
    \begin{itemize}[\qquad]
        \item (neural-network) quantile regressors (\textbf{QR}) from~\cite{koenker2001quantile,cqr},
        \item quantile regression forests (\textbf{QRF}) from~\cite{meinshausen2006quantile},
        \item mean-variance estimators (\textbf{MV}) from~\cite{nix1994estimating},
        \item mean-variance ensembles (\textbf{MVE}
        ) from~\cite{kendallgal}.
    \end{itemize}
    Each of these methods comes in different flavours when augmented with conformal prediction. All of them have a baseline performance (no conformal prediction), a marginal CP incarnation giving rise to three models (residual score~\eqref{residual_score}, interval score~\eqref{interval_score} and normalized score~\eqref{normalized_score}) and a Mondrian CP incarnation, again giving rise to three models. The three variants are, respectively, denoted by PointCP, IntCP and NCP, whereas the marginal variants are further indicated by the prefix `m'. For the conditional performance, this results in seven different options. The aim of this experimental part is to analyse whether the results from the previous section hold, i.e.~to check when the marginal models, the normalized variant in particular, give the same performance as the conditional (Mondrian) ones. For more information about the models, choices of architecture and further hyperparameter choices, see Appendix~\ref{appendix:models}.

    All experimental results were obtained by evaluating the models on 10 different train/test-splits. The test set always contained 20\% of the data. To obtain a calibration set, the training set was further split in half. The significance level was fixed at $\alpha=0.1$.

\subsection{Real data}

    Most of the data sets were obtained from the UCI repository~\cite{Dua2019}. Specific references are given in Table~\ref{tab:datasets}. This table also shows the number of data points and (used) features, together with the skewness and (Pearson) kurtosis of the response variable. All data sets were standardized (both features and target variables) before training. For the taxonomy function $\kappa:\mathcal{X}\times\mathbb{R}\rightarrow\mathcal{C}$, equal-frequency binning of the variance estimates with $n=3$ bins was chosen, as in the synthetic case. Although these data sets all have different characteristics, e.g.~dimensionality, sparsity, count data vs.~continuous data, etc., they are treated equally. The experimental outcomes could be interpreted in light of these differences, but this would require more insight in the underlying data-generating mechanism.

    \begin{table}[t!]
        \caption{Overview of the data sets.}
        \label{tab:datasets}
        \centering
        \renewcommand{\arraystretch}{1.2}
        \begin{tabular}{ccccc}
            \toprule
            Name&\# samples&\# features&Skewness\ /\ Kurtosis&Source\\
            \midrule
            \texttt{concrete}&1030&8&0.42\ /\ 2.68&\cite{yeh1998modeling}\\
            \texttt{turbine}&9568&4&0.31\ /\ 1.95&\cite{kaya2012local}\\
            \texttt{puma32H}&8192&32&0.02\ /\ 3.04&\cite{corke1996robotics}\\
            \texttt{residential}&372&105&1.26\ /\ 5.15&\cite{rafiei2016novel,misc_residential_building_data_set_437}\\
            \texttt{crime2}&1994&123&1.52\ /\ 4.83&\cite{redmond2002data,misc_communities_and_crime_183}\\
            \texttt{star}&2161&39&0.29\ /\ 2.63&\cite{star_data}\\
            \bottomrule
        \end{tabular}
    \end{table}

    The coverage of the prediction intervals of different models on the data sets from Table~\ref{tab:datasets} are shown in Figs.~\ref{fig:cov_conditional} and~\ref{fig:cov_conditional2}. (For figures of the PI widths and marginal performance for all data sets, see Appendix~\ref{appendix:real_data}.) It should be immediately clear that the Mondrian conformal prediction methods, indicated by the labels `PointCP', `IntCP' and `NCP', have the most stable and desirable behaviour. For these methods the coverage is centered around the target confidence level of 90\% (corresponding to the predetermined significance level of $\alpha=0.1$), whereas for the baseline and marginal conformal prediction methods, the coverage can fluctuate heavily, lying either far above or far below 90\%. Of course, this is entirely to be expected, since only the Mondrian approach satisfies the strict conditional coverage guarantees of Theorem~\ref{conditional_validity}.

    Another feature of these figures is that the deviations of the marginal models are not always in the same direction. Comparing the different subplots, it can be seen that whereas some models show undercoverage on one data set, they exhibit overcoverage on the other. The same occurs among the different methods on a single data set. If the true variance were known, as in Figs.~\ref{fig:variance_cdf} and~\ref{fig:variance_comparison} in the previous section, an in-depth analysis could be made. However, in contrast to estimates of the true response, where one can use metrics such as the MSE or $R^2$ to quantify deviations from the truth, no analogous approaches exist for the residual variance. For higher conditional moments, multiple measurements for the same feature tuple are required to get a good estimate. Any uncertainty-dependent taxonomy function will, therefore, also lead to a wrong decomposition of the instance space. Moreover, there is in general no way to determine where the lack of validity stems from: bad estimates of the prediction intervals or incorrect taxonomy classes ({\em cf.}~the distinction between misspecification and contamination in Section~\ref{section:synthetic}).

%% file: Appendix.tex
\section{Model architectures}\label{appendix:models}

    All neural networks were constructed using the default implementations from \texttt{PyTorch}~\cite{pytorch}. Their general architecture was fixed. As in~\cite{cqr,dewolf2023valid} the Adam optimizer was used for weight optimization with a fixed learning rate of $5\times10^{-4}$. The number of epochs was fixed at 300. All networks contained three hidden layers, each with 64 neurons. The activation functions after all layers, except the final one, were of the ReLU type, while the activation function at the output node was simply a linear function. For regularization, all models, except the normalizing flows which used the implementation from Wehenkel \textit{et al.}~\cite{wehenkel2019unconstrained}, also had a Dropout layer~\cite{srivastava2014dropout} before each hidden layer with dropout probability fixed at $0.2$. For a general introduction to most of the models, see Dewolf \textit{et al.}~\cite{dewolf2023valid}. For completeness, each of the models from the list above, is explained here in a little more detail:
    \begin{enumerate}
        \item \textit{Quantile regression}:
        Since, except for the baseline version, the models are augmented by conformal prediction, the models do not have to be trained at extreme significance levels, where data might be scarce. Therefore, a `softening factor' as in~\cite{cqr,sesia2020comparison} was adopted. In this paper, the value was fixed at $w=2$.
        \begin{enumerate}[(i)]
            \item[(a)] \textit{Neural networks}: A neural network with two (or three if the median is used as a point estimate) outputs is trained using the pinball loss~\cite{cqr,dewolf2023valid}. If the mean is preferred over the median, an ordinary MSE term can be added to the loss function.
            \item[(b)] \textit{Random forests}: Instead of taking the mean of samples in the leaf nodes of the trees, all the samples are retained to construct a full distribution function~\cite{meinshausen2006quantile}. The default implementation from Nelson~\cite{jnelson18_2023} was used.
        \end{enumerate}
        \item \textit{Mean-variance estimation}:
        As for quantile regression, a neural network with two outputs is used. Up to a logarithmic transformation for positivity and numerical stability, these yield the mean and the variance. The model is then trained by optimizing the (log)likelihood of a normal distribution with the given mean and variance.
        \item \textit{(Dropout) Ensemble estimation}:
        Using Dropout at test time allows to create an ensemble without having to actually (re)train a network multiple times. This considerably speeds up the training process. In this paper, the mean-variance approach from Kendall \textit{et al.}~\cite{kendallgal} with an ensemble of 50 Monte Carlo samples was used, i.e.~50 mean-variance models were obtained. The Dropout-ensemble constitutes a mixture of Gaussian distributions with parameters $\{(\widehat{\mu}_i,\widehat{\sigma}_i)\}_{i\leq50}$. Aggregating these estimates gives a total mean and total variance. The problem of finding the quantiles is simplified by approximating the mixture model by a Gaussian distribution with the same mean and variance. This allows to find the prediction intervals in the standard way.
    \end{enumerate}

\section{Real data}\label{appendix:real_data}
\begin{figure}[t!]
        \centering
        \includegraphics[width = \linewidth]{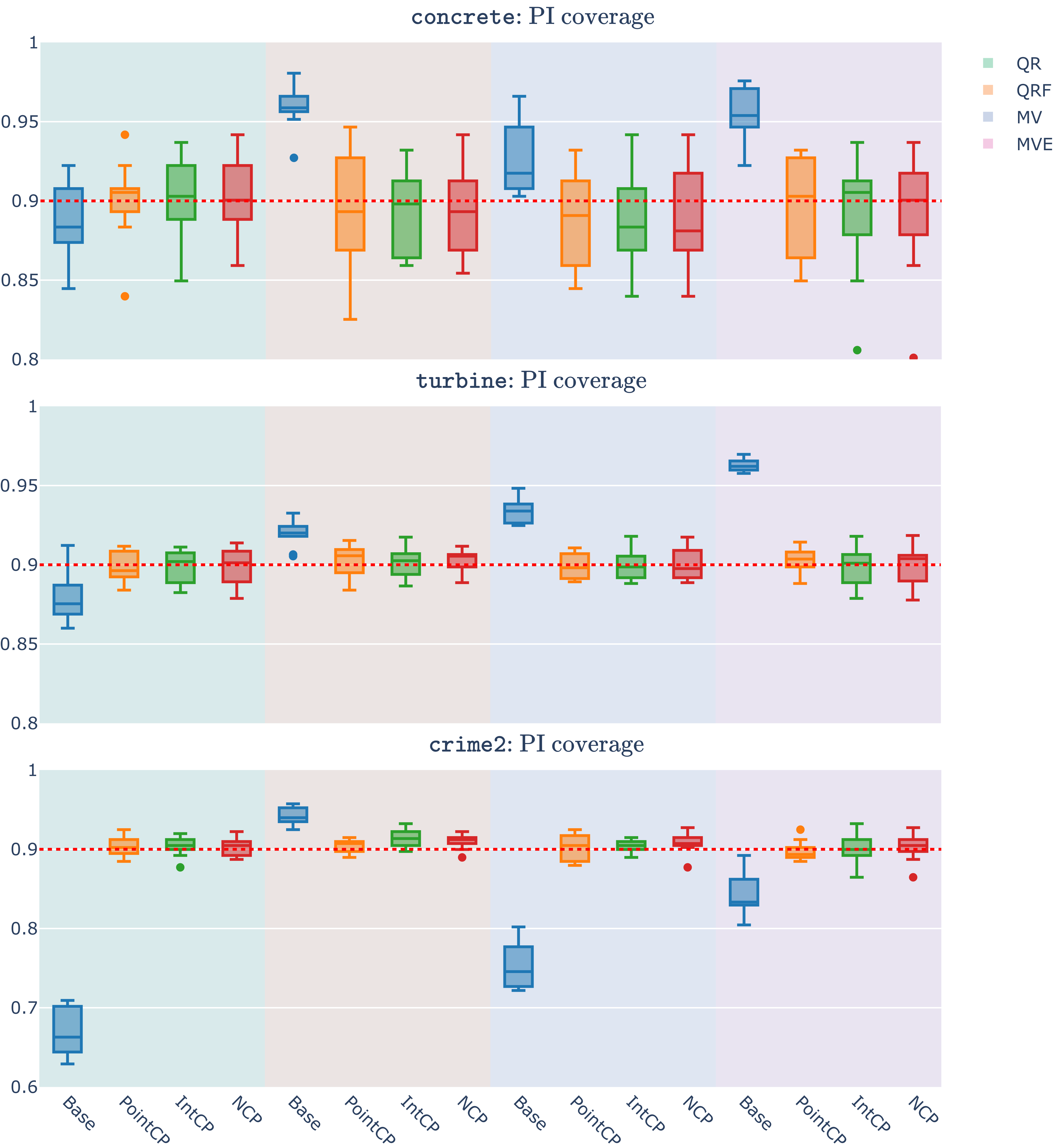}
        \caption{Marginal coverage at significance level $\alpha=0.1$ for the \texttt{concrete}, \texttt{turbine} and \texttt{crime2} data sets. The coloured columns indicate the different estimators (from left to right): quantile regression, quantile regression forest, mean-variance estimator and mean-variance ensemble. For every model, a baseline result and three nonconformity measures are shown (from left to right): residual, interval and $\widehat{\sigma}$-normalized nonconformity measures.}
        \label{fig:cov}
    \end{figure}

\begin{figure}[t!]
        \centering
        \includegraphics[width = \linewidth]{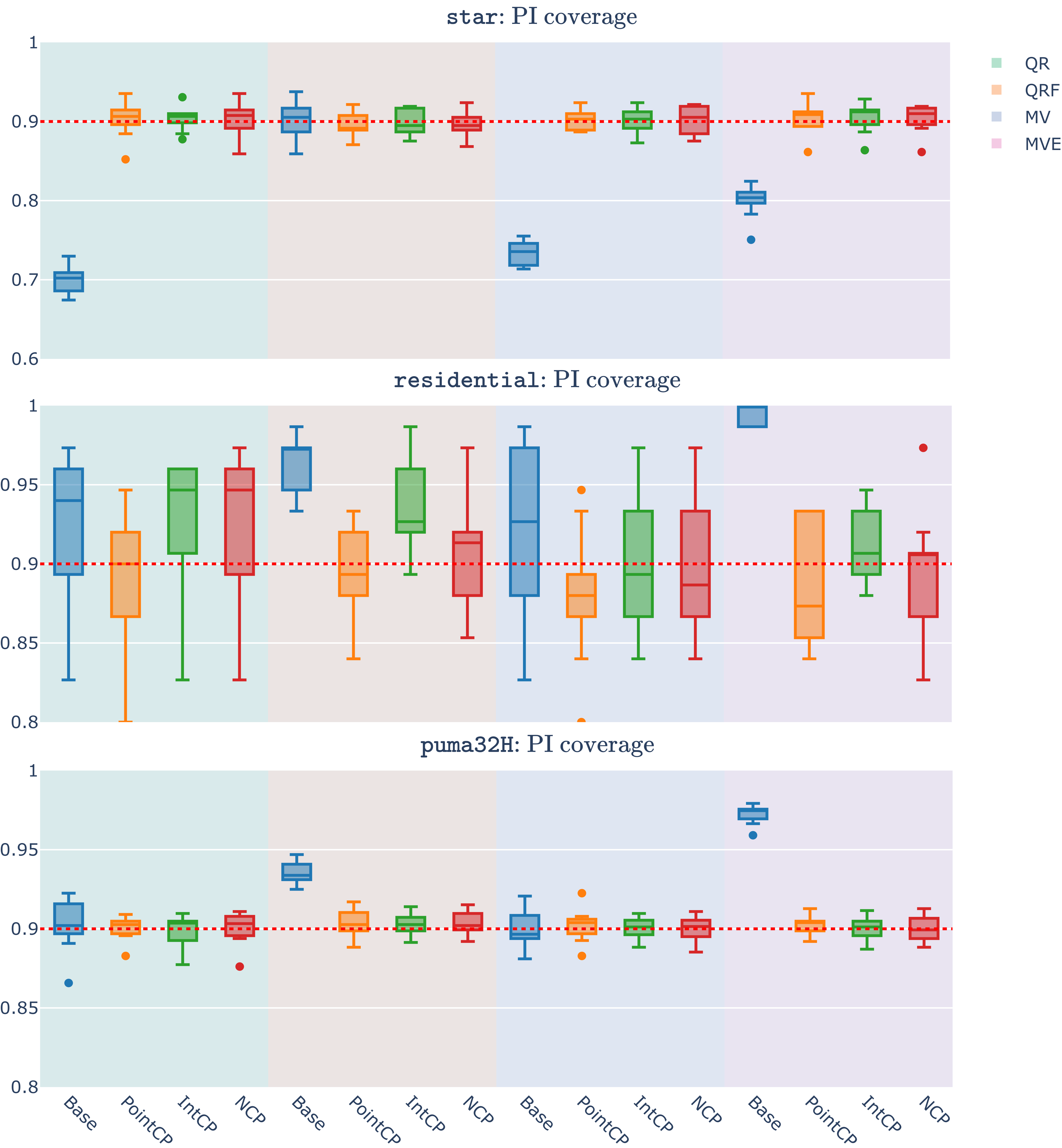}
        \caption{Marginal coverage at significance level $\alpha=0.1$ for the \texttt{star}, \texttt{residential} and \texttt{puma32H} data sets. The coloured columns indicate the different estimators  (from left to right): quantile regression, quantile regression forest, mean-variance estimator and mean-variance ensemble. For every model, a baseline result and three nonconformity measures are shown (from left to right): residual, interval and $\widehat{\sigma}$-normalized nonconformity measures.}
        \label{fig:cov2}
    \end{figure}

\begin{figure}[t!]
        \centering
        \includegraphics[width = \linewidth]{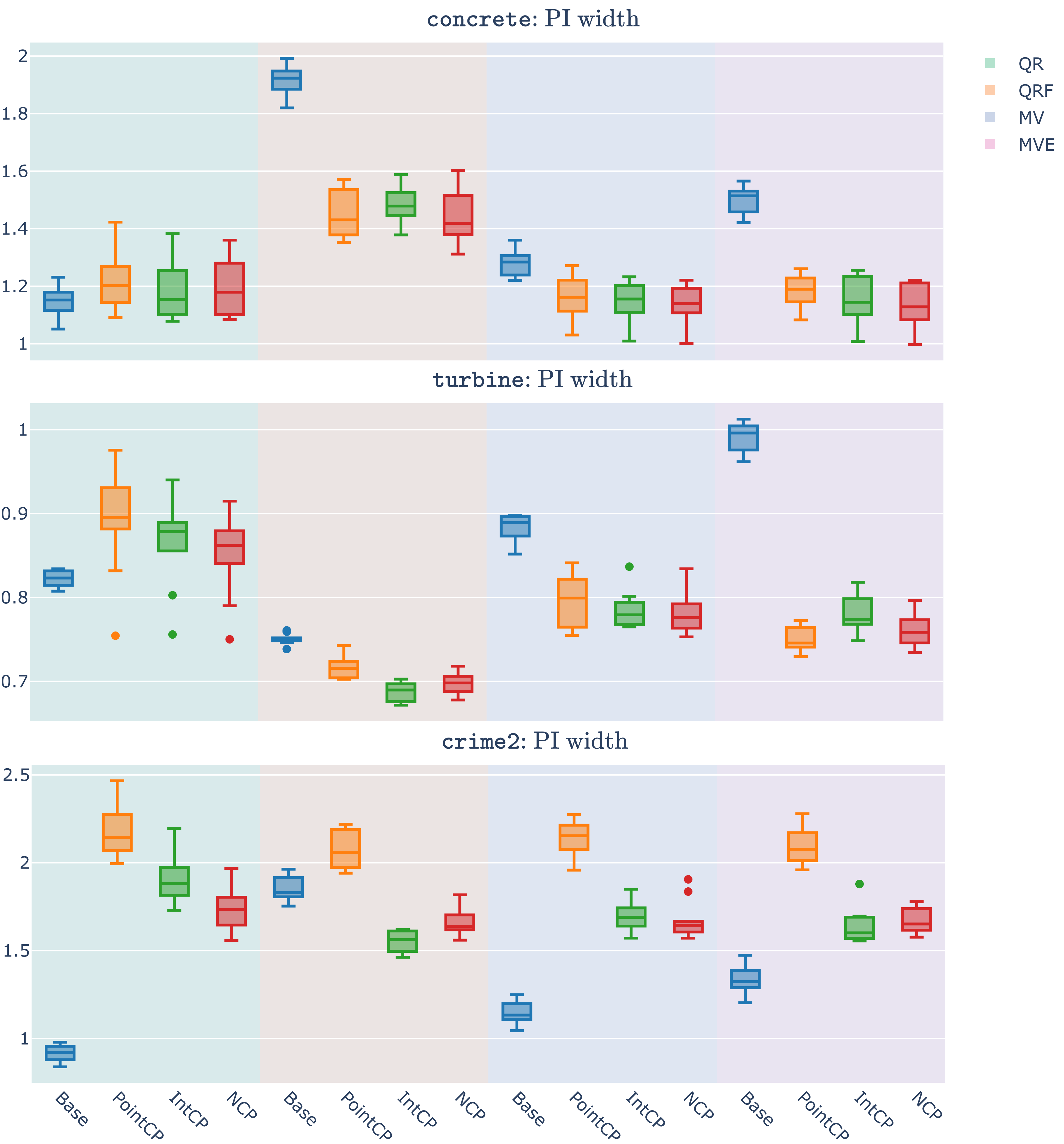}
        \caption{Marginal PI widths at significance level $\alpha=0.1$ for the \texttt{concrete}, \texttt{turbine} and \texttt{crime2} data sets. The coloured columns indicate the different estimators  (from left to right): quantile regression, quantile regression forest, mean-variance estimator and mean-variance ensemble. For every model, a baseline result and three nonconformity measures are shown (from left to right): residual, interval and $\widehat{\sigma}$-normalized nonconformity measures.}
        \label{fig:width}
    \end{figure}

\begin{figure}[t!]
        \centering
        \includegraphics[width = \linewidth]{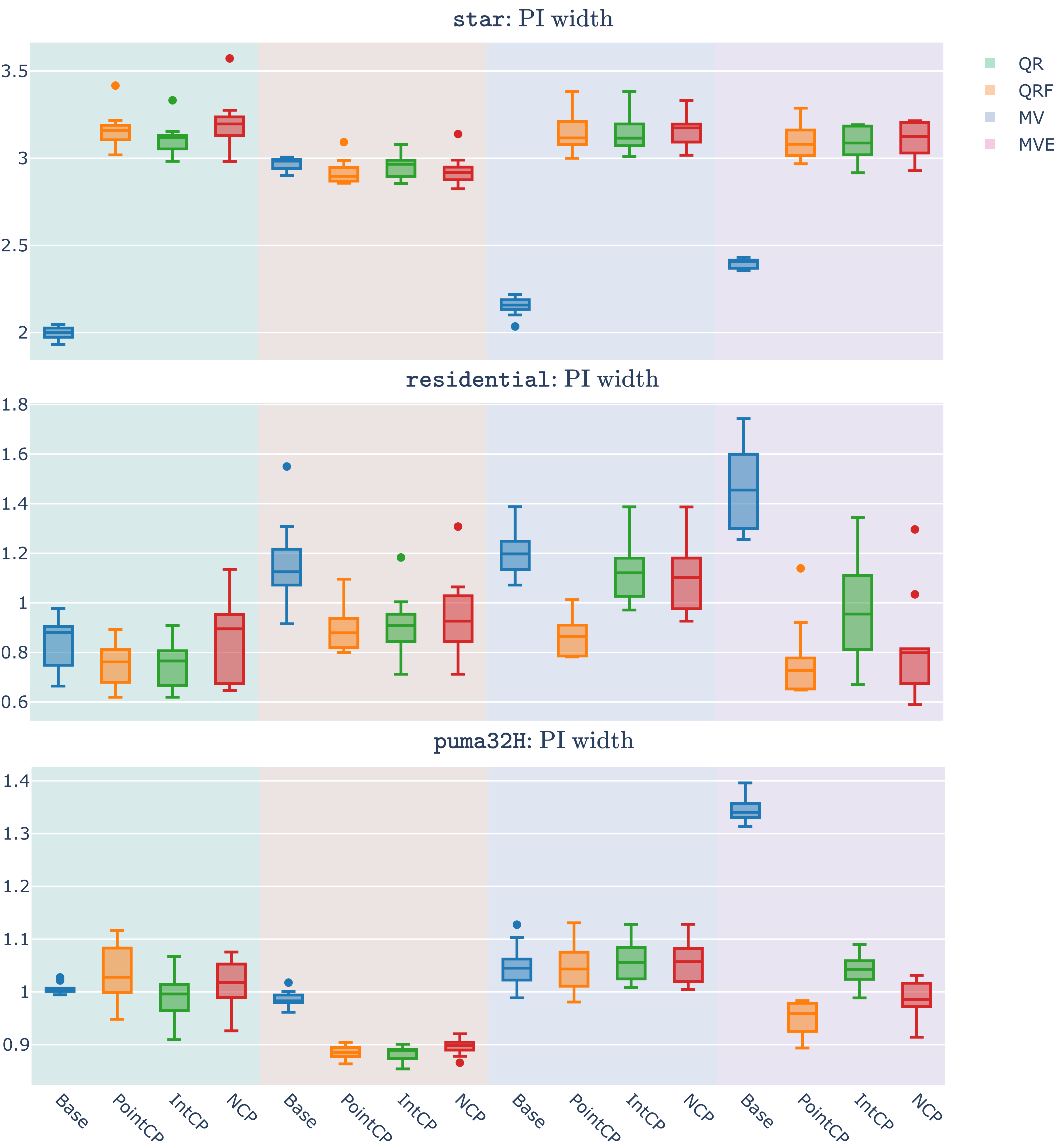}
        \caption{Marginal PI widths at significance level $\alpha=0.1$ for the \texttt{star}, \texttt{residential} and \texttt{puma32H} data sets. The coloured columns indicate the different estimators  (from left to right): quantile regression, quantile regression forest, mean-variance estimator and mean-variance ensemble. For every model, a baseline result and three nonconformity measures are shown (from left to right): residual, interval and $\widehat{\sigma}$-normalized nonconformity measures.}
        \label{fig:width2}
    \end{figure}

    Figures~\ref{fig:cov}--\ref{fig:width2} show the marginal coverage and width, i.e.~without conditioning by a taxonomy function, of the prediction intervals constructed by all models from Section~\ref{section:models}: QR, QRF, MV and MVE.

\begin{figure}[t!]
        \centering
        \includegraphics[width = .8\linewidth]{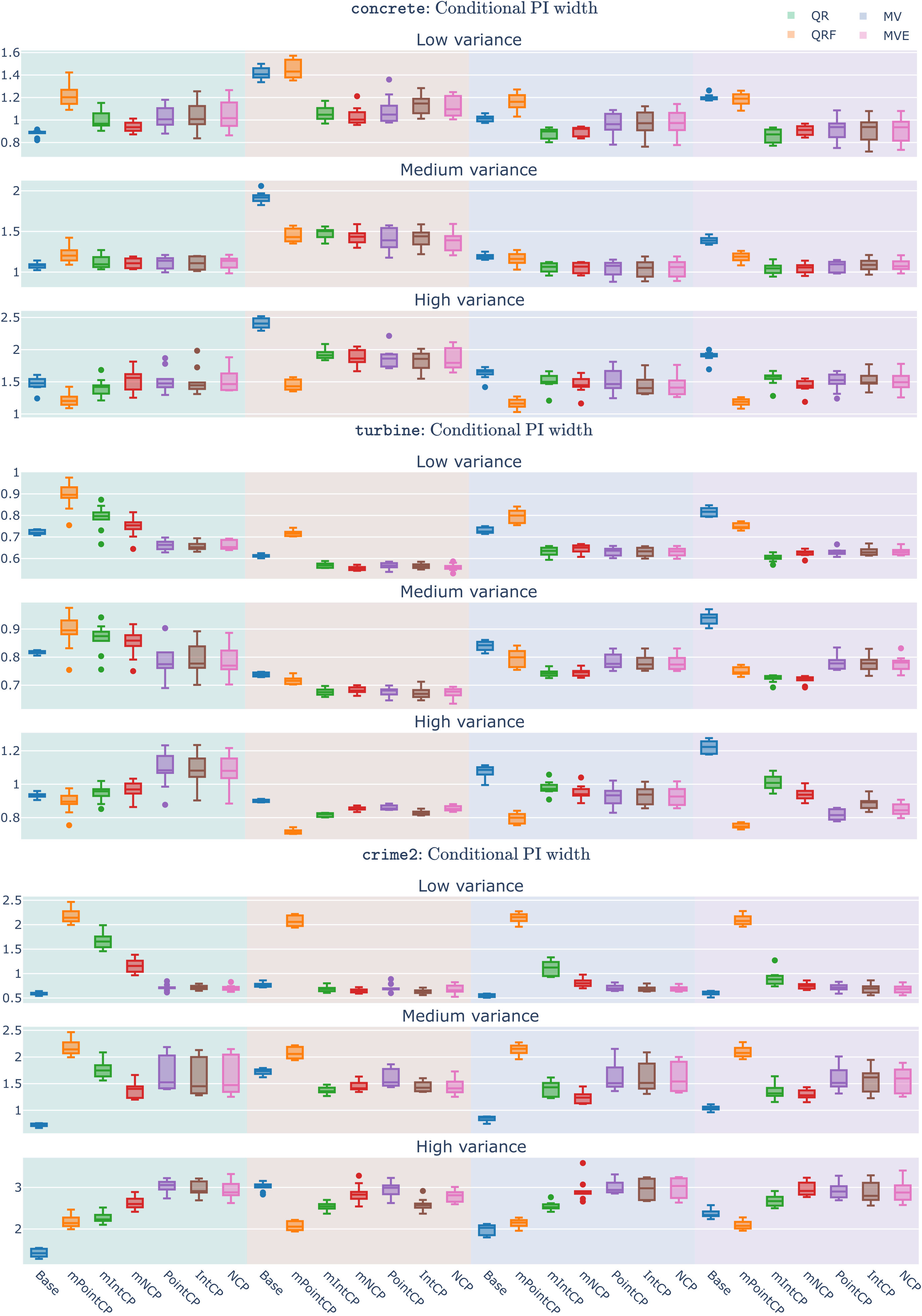}
        \caption{Conditional PI widths at significance level $\alpha=0.1$ for the \texttt{concrete}, \texttt{turbine} and \texttt{crime2} data sets. The data is divided in three folds based on equal-frequency binning of the estimated variance. The coloured columns indicate the different estimators  (from left to right): quantile regression, quantile regression forest, mean-variance estimator and mean-variance ensemble. For every model, a baseline result and six nonconformity measures are shown (from left to right): residual, interval and $\widehat{\sigma}$-normalized nonconformity measures and their Mondrian counterparts.}
        \label{fig:width_conditional}
    \end{figure}

\begin{figure}[t!]
        \centering
        \includegraphics[width = .8\linewidth]{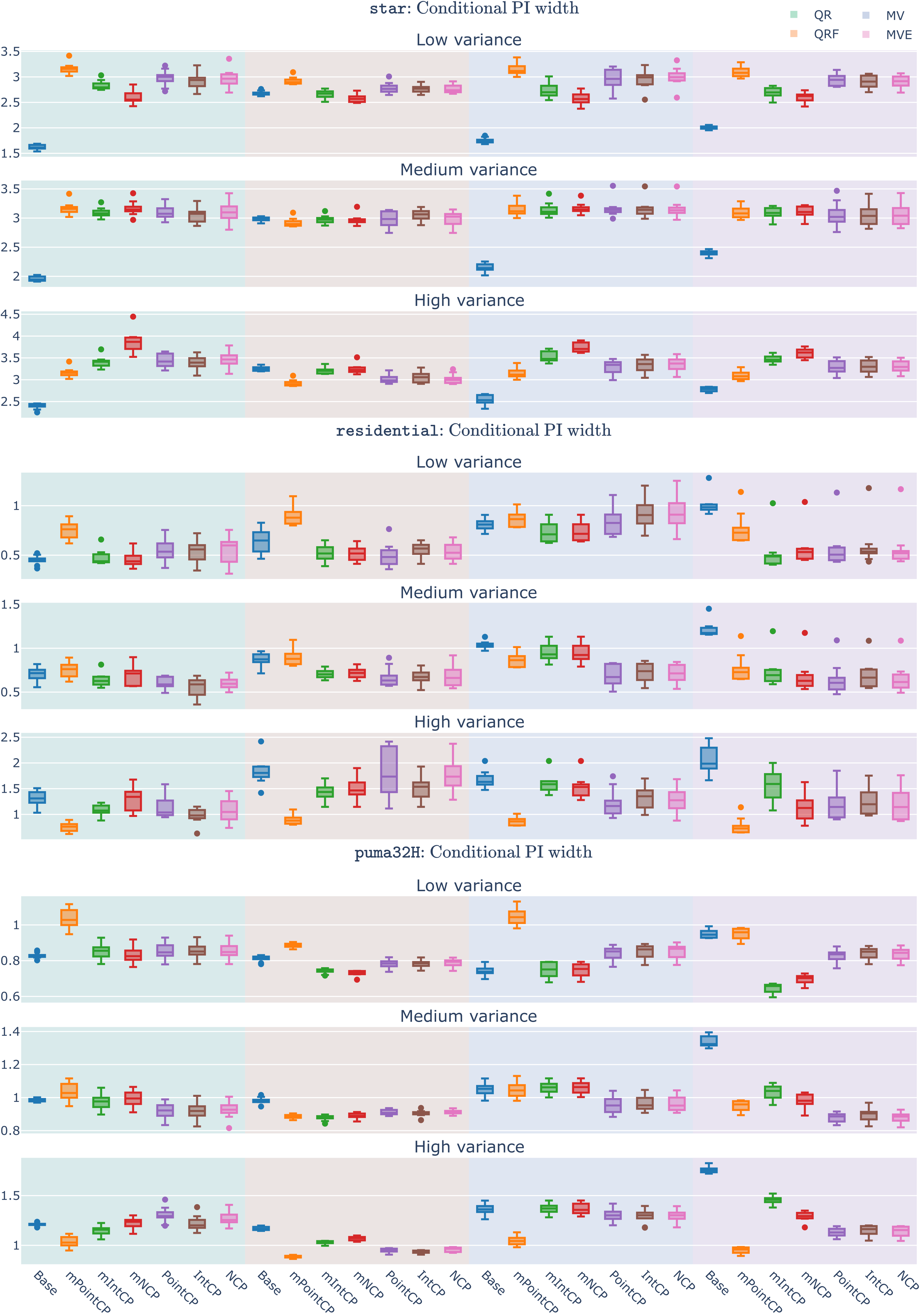}
        \caption{Conditional PI widths at significance level $\alpha=0.1$ for the \texttt{star}, \texttt{residential} and \texttt{puma32H} data sets. The data is divided in three folds based on equal-frequency binning of the estimated variances. The coloured columns indicate the different estimators  (from left to right): quantile regression, quantile regression forest, mean-variance estimator and mean-variance ensemble. For every model, a baseline result and six nonconformity measures are shown (from left to right): residual, interval and $\widehat{\sigma}$-normalized nonconformity measures and their Mondrian counterparts.}
        \label{fig:width_conditional2}
    \end{figure}

    Figures~\ref{fig:width_conditional} and~\ref{fig:width_conditional2} show the conditional coverage and width, using equal-frequency binning of the variance estimates ($n=3$), of the prediction intervals constructed by all models from Section~\ref{section:models}.

